\documentclass[10pt]{article}

\usepackage{epsfig}
\usepackage{amsmath}
\usepackage{amssymb}
\usepackage{color}
\usepackage{url}

\usepackage{verbatim} 
\usepackage{epstopdf} 
\usepackage{graphicx} 
\usepackage[font=small]{caption} 

\usepackage{afterpage}

\usepackage{caption}
\usepackage{subcaption}
\usepackage{multicol}

\usepackage{multirow}

\usepackage{setspace}

\usepackage[numbers]{natbib}

\newcommand{\commentout}[1]{}

\newcommand{\R}{\mathbb{R}}

\renewcommand{\arraystretch}{1.2}

\begin{document}

\title{Spatio-Temporal Surrogates for Interaction of a Jet with High
  Explosives: Part I - Analysis with a Small Sample Size}

\author{%
Chandrika Kamath, Juliette S. Franzman, and Brian H. Daub\\[0.5em]
{\small\begin{minipage}{\linewidth}\begin{center}
\begin{tabular}{c}
Lawrence Livermore National Laboratory \\
7000 East Avenue, Livermore, CA 94551, USA\\
\url{kamath2, franzman1,daub1@llnl.gov}\\
\hspace*{0.8in}
\end{tabular}
\end{center}\end{minipage}}
}

\date{9 June 2023}
\maketitle

\begin{abstract}

  Computer simulations, especially of complex phenomena, can be
  expensive, requiring high-performance computing resources. Often, to
  understand a phenomenon, multiple simulations are run, each with a
  different set of simulation input parameters.  These data are then
  used to create an interpolant, or surrogate, relating the simulation
  outputs to the corresponding inputs.  When the inputs and outputs
  are scalars, a simple machine learning model can suffice. However,
  when the simulation outputs are vector valued, available at
  locations in two or three spatial dimensions, often with a temporal
  component, creating a surrogate is more challenging.  In this
  report, we use a two-dimensional problem of a jet interacting with
  high explosives to understand how we can build high-quality
  surrogates. The characteristics of our data set are unique - the
  vector-valued outputs from each simulation are available at over two
  million spatial locations; each simulation is run for a relatively
  small number of time steps; the size of the computational domain
  varies with each simulation; and resource constraints limit the
  number of simulations we can run.  We show how we analyze these
  extremely large data-sets, set the parameters for the algorithms used
  in the analysis, and use simple ways to improve the accuracy of the
  spatio-temporal surrogates without substantially increasing the
  number of simulations required.

\end{abstract}

\pagebreak
\begin{spacing}{0.7}
\tableofcontents
\end{spacing}

\pagebreak

%
\section{Introduction}
%

Computer simulations are increasingly being used in science and
engineering applications. However, it can be time consuming to run
these simulations for a given set of input parameters, especially when
the problem being modeled is complex and requires high-performance
computing resources.  Surrogates, often based on a machine learning
model~\cite{kamath2018:regression}, are used to provide a fast, but
approximate, alternative that relates the simulation outputs to the
corresponding inputs. Such surrogates are relatively easy to create
when the simulation inputs and outputs are both scalars. However, when
the output is in two or three spatial dimensions and varies with time,
relating these spatio-temporal outputs to the scalar inputs becomes
more challenging. If, in addition, we are constrained by time or
computer resources to generate data for only a small number of
simulations, building a surrogate that is accurate, becomes non-trivial.

We describe our work in creating surrogates for a problem in two
spatial dimensions, plus time, in two reports. In this first report,
we discuss the applications aspect of our work, focusing on issues
related to the small number of simulations in our data set.  We want
to understand whether it is possible to build an accurate, predictive
surrogate model under these conditions and to identify simple ways to
improve the accuracy of these models.  In a companion
report~\cite{kamath2023:stmclus}, we discuss in detail how we made our
solution approach tractable, despite the large size of the spatial
data generated at each time step of the simulation.

We start this report by describing the problem considered, namely, the
interaction of a jet with high explosives, and the two-dimensional
outputs generated by the simulations (Section~\ref{sec:data_desc}).
We then discuss the unique aspects of our problem and place it in the
context of related work (Section~\ref{sec:related}). We describe how
we address these unique challenges, focusing on ways in which we can
improve the accuracy of the surrogate models
(Section~\ref{sec:approach}).  Using several test cases, we explore
how to set various parameters in the algorithms used in our solution
approach and the metrics to evaluate the predictions of the surrogate
models (Section~\ref{sec:results}).  Finally, we conclude this report
with the lessons learned in building accurate spatio-temporal
surrogates for a limited number of simulations, each of which
generates a large amount of data (Section~\ref{sec:conc}).

%
\section{Description of the problem and data}
\label{sec:data_desc}
%

\begin{figure}[!b]
\centering
\begin{tabular}{c}
\includegraphics[trim = 0.0cm 5.5cm 0.0cm 6.5cm, clip = true,width=0.65\textwidth]{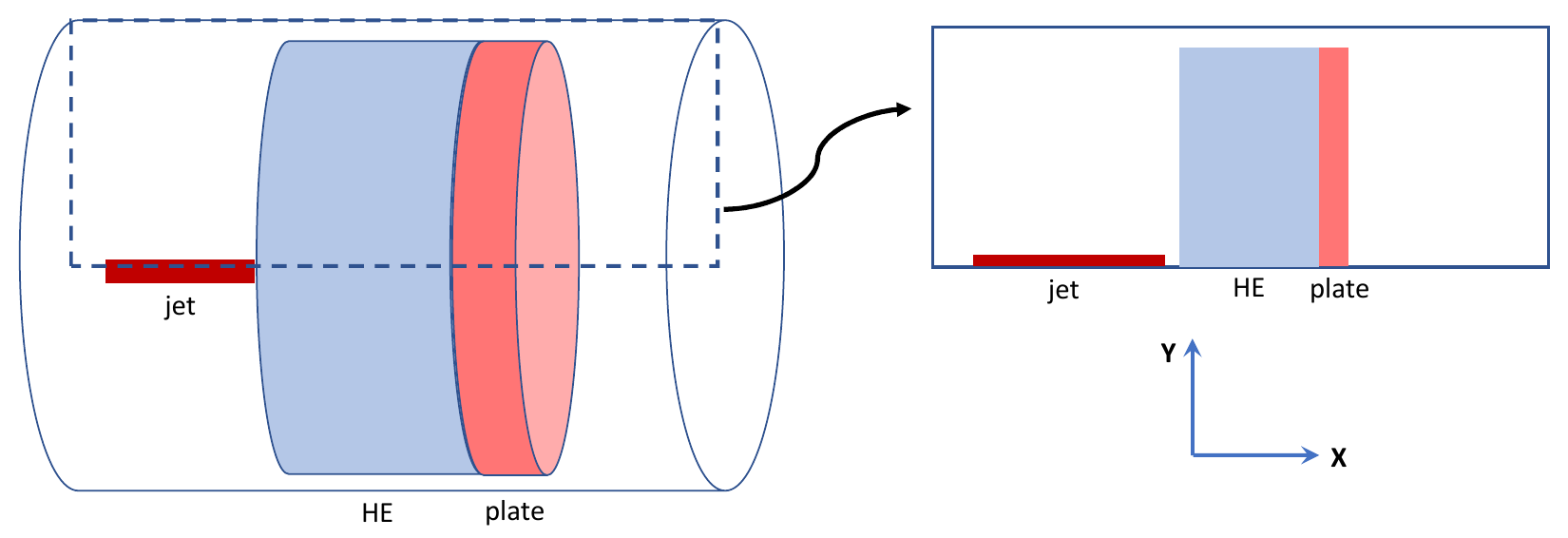}\\
\end{tabular}
\caption{A schematic of the problem being simulated. On the left is
  the horizontal cylinder in three-dimensions, showing the plate in
  pink, to the left of which is the HE in blue. The jet, in red,
  enters the HE from the left. As the problem is radially symmetric
  about the axis of the cylinder, we need to simulate only the
  two-dimensional region shown by a dotted rectangle on the left and
  schematically on the right. }
\label{fig:schematic}
\end{figure}

We illustrate our ideas on building accurate spatio-temporal
surrogates using simulation output from a problem describing the
interaction of a jet with high explosives (HE). The domain of the
problem is a right cylinder with its axis oriented horizontally as
shown on the left in Figure~\ref{fig:schematic}. There is a steel
plate, 1cm thick, near the right end of the cylinder, with the LX14
high explosive to the left of the plate. Both the plate and the HE
have a fixed radius of 10cm. A copper jet aligned along the center
line of the cylinder, enters the HE from the left. The simulation
models what happens as the jet moves through the HE and the plate.
The jet is modeled initially as uniform cylinder. It is 10 cm in
length with a varying radius. The jet tip velocity is specified as an
input parameter; a linearly-varying velocity profile is applied to
the remainder of the cylinder that represents the jet to approximate a
stretching metal jet.  As the problem is radially symmetric about the
axis of the cylinder, only the two-dimensional region shown by a
dotted rectangle on the left in Figure~\ref{fig:schematic}, and
schematically on the right, is simulated.

There are three input parameters for the simulation: the {\it radius}
of the jet, the {\it length} of the HE to be traversed by
the jet, and the {\it tip velocity} of the jet in the positive $x$
direction.  At each time step, the simulation outputs variables of
interest, such as mass and momentum, at different points on a grid in the two
dimensional region.  By running the simulations at select values of
these input parameters, and collecting the output at different time
steps for each simulation, we can create a data set that could be used
to build a surrogate model to predict the output at a new set of input
parameters and a given time step. We are interested in determining for
example, whether the plate breaks; what is the final position of the
plate; and, if the plate breaks, what is the velocity of the jet tip as it
comes out on the other side of plate.

\begin{figure}[!htb]
\centering
\begin{tabular}{cc}
\includegraphics[width=0.5\textwidth]{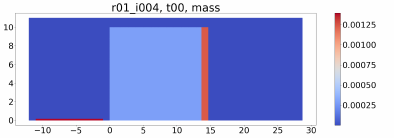} &
\includegraphics[width=0.5\textwidth]{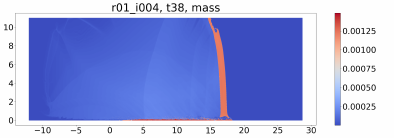} \\
\includegraphics[width=0.5\textwidth]{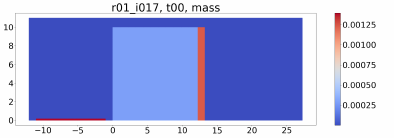} & 
\includegraphics[width=0.5\textwidth]{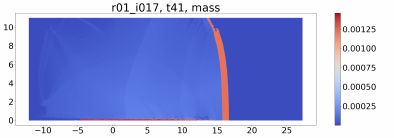} \\
\includegraphics[width=0.5\textwidth]{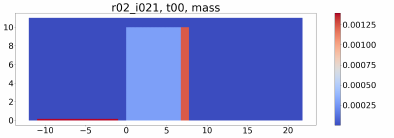} & 
\includegraphics[width=0.5\textwidth]{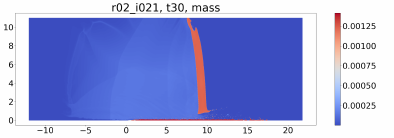} \\
\includegraphics[width=0.5\textwidth]{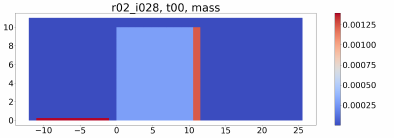} & 
\includegraphics[width=0.5\textwidth]{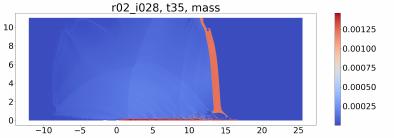} \\
\end{tabular}
\caption{The variable {\it mass} at the first time step (left column)
  and last time step (right column) for the four example simulations
  in Table~\ref{tab:sample_params}. From top to bottom, simulations
  with keys r01\_i004, r01\_i017, r02\_i021, r02\_i028, illustrate an
  {\it almost break}, {\it no break}, {\it break}, and {\it break}
  case, respectively.  The data shown are before the pre-processing
  steps described in Section~\ref{sec:preprocessing}. The vertical red
  region represents the plate, the light blue represents the high
  explosive (HE), the horizontal red region at the bottom to the left
  of the HE at the first time step is the jet, and the dark blue
  region is air. The right column shows the last time step of each
  simulation and the effect on the plate as the jet moves to the
  right, through the HE, and potentially through the plate. The range
  of x values is different for different simulations, while the range
  of y values is the same. The HE starts at $x=0$ in each plot. Note
  that even when the plate does not break (second row), the plate has
  moved to the right from its original position.}
\label{fig:sample_snaps_nmass}
\end{figure}

To illustrate the instances in our data set, we use four simulations
whose parameters are listed in Table~\ref{tab:sample_params}.
Figure~\ref{fig:sample_snaps_nmass} shows the output variable, {\it
  mass}, at the first and last time steps for these four example
simulations.  As explained earlier, we have simplified the
three-dimensional problem by assuming radial symmetry around the axis
of the cylinder, so the output from the simulation is shown as two
dimensional images, 
with the axis of the cylinder shown at the bottom, that is, at $y =
0$.  The domain extent in $x$ (along the length of the cylinder)
varies as the length of the HE varies across simulations; however, the
domain in $y$ ranges from 0 to 11cm for all simulations.

\renewcommand{\arraystretch}{1.2}
\begin{table}[htb]
  \begin{center}
    \begin{tabular}{|l|l|l|l|l|l|l|}
      \hline
      Simulation key & jet radius & HE-length & jet tip velocity & \texttt{\#} time & \texttt{\#}grid & outcome \\
                     & (cm)   & (cm)      & (cm/$\mu$sec) & steps & points & \\
      \hline
      r01\_i004 & 0.15 & 13.67 & 0.894 & 38 & 2,859,387 & almost break \\
      r01\_i017 & 0.17 & 12.24 & 0.648 & 41 & 2,759,181 & no break \\
      r02\_i021 & 0.14 & 6.77 & 0.914  & 30 & 2,374,179 & break \\
      r02\_i028 & 0.23 & 10.54 & 0.843 & 35 & 2,639,637 & break \\
      \hline
    \end{tabular}
  \end{center}
  \captionof{table}{Input parameters for the four example simulations
    shown in Figure~\ref{fig:sample_snaps_nmass}. Note the very large 
    number of grid points (over two million) at which variables of 
    interest are output at each time step in a simulation.}
  \label{tab:sample_params}
\end{table}

In Figure~\ref{fig:sample_snaps_nmass}, the vertical plate, shown in
red, is stationary at time $t = 0$. To the left of the plate is the HE
shown in light blue.  The jet in shown in red at the bottom of the
domain to the left of the HE; it is quite thin relative to the radius
of the cylinder, and is barely visible in the images.  As the
simulation evolves, the jet moves to the right, through the HE, which
expands, pushing the plate to the right. At late time, depending on
the simulation input parameters, the plate could:
\begin{itemize}

\item{\bf break}, with the jet going through the plate and 
  coming out clearly on the other side;

\item{\bf almost break}, with the jet either going completely through
  the plate but barely coming out the other side or the jet going
  almost all the way through the plate, leaving it barely connected at
  the bottom;

\item{\bf not break}, with the plate remaining attached, either
  partially or completely, at the bottom. The plate could have moved
  from its original position at time $t =0$.

\end{itemize}  
We used the last two time steps in each simulation to assign one of
these three class labels to the simulation. This label was
not used in building the surrogate; it was used only to ensure we had
a good coverage of the design space. We selected the four example
simulations in Figure~\ref{fig:sample_snaps_nmass} to illustrate these
three cases.

In our simulations, the output at each time step consists of the
values of variables of interest that are generated at grid points in
the two-dimensional rectangular domain. These grid points are on a
regular grid, with $\Delta x = \Delta y = 0.0125$cm.  There are three
variables output: {\it mass, x-momentum, and y-momentum}; the latter
two are shown later in Appendices~\ref{sec:appendix1}
and~\ref{sec:appendix2}, respectively.  The values of these variables
are defined at the center of the cell formed by four nearby grid
points. Thus the data appear as an image, with regularly spaced
pixels. However, in general, the grid points in a simulation need not
be on a regular grid; they could form an unstructured grid, as in a
finite element mesh, or a locally structured grid, as in an Adaptive
Mesh refinement (AMR) mesh.  As a result, unlike an image, most output
from simulations also include the $(x,y)$ coordinates of the grid
points.  In our work, we retain this association of the coordinates
with the grid points as it enables us to extract sub-domains from the
larger domain for processing.

Each simulation is run for a fixed number of time steps which is
determined as $( \lfloor (\text{HE-length}/\text{jet-tip-velocity} \rfloor ) + 23
)$, with the output generated at each time step. As both HE-length and
jet tip velocity vary with the simulation, the number of time steps also varies
across simulations. At early time, as the jet starts to move through
the HE, there is little of interest in the simulation output. Once the
jet is partway through the HE, as indicated by the first term in the
equation above, it starts to influence the location of the plate,
until 23 $\mu$sec later, it is expected that we should know the final
status of the plate. In our work, we consider all the time steps in
the analysis; an alternative would be to consider only the later 23
time steps.

The output data for a variable at a time step in a simulation is
referred to as a {\it snapshot}, so named as it is a snapshot of the
evolution of the simulation at a particular point in time.  For the
problem considered in this report, we generated the data set by
identifying sample parameter values in the three-dimensional input
space and running the corresponding simulations for the specified
number of time steps. For each of the three output variables, the data
set consists of the snapshots across all time steps of all
simulations.

Our eventual goal in building the spatio-temporal surrogates for the
jet-HE interaction problem is to predict what happens to the jet and
the plate at late time, specifically, does the jet go through the
plate, what is the velocity of the jet tip when it goes through the
plate, and what is the location of the plate at late time. However, in
this initial study, we limit the scope of the work and explore options
for creating an accurate surrogate when we have a small number of
simulations.  Specifically, we consider how to process large data
sets, how to set parameters for the algorithms used in our solution
approach, and how to improve the accuracy of the surrogates without
substantially increasing the number of simulations required.  We
evaluate our ideas using a qualitative comparison of predicted outputs
for seven test snapshots.

%
\subsection{Challenges to the analysis}
\label{sec:challenges}
%

There are two main challenges to building spatio-temporal surrogates
for our problem:

\begin{itemize}

\item The first is how do we build a surrogate that is {\it accurate}?
  The predictive accuracy depends on two factors. The first is
  the quality of the training data. Simulating the jet-HE interaction
  is resource-intensive, requiring multiple processors of a
  high-performance computing system.  This limits the number of sample
  points we can run in the three-dimensional input parameter space.
  These points have to be selected carefully; this is difficult as we
  do not know {\it a priori} the outcome of running the simulation at
  a specific sample point, which in turn implies that we do not know
  the range of parameter values to use. The second factor influencing
  the accuracy of the surrogate is the choice of the model used.  We
  want to predict a two-dimensional output, given only four scalar
  inputs - the three simulation input parameters and a time step. A
  traditional machine learning model, where we predict scalar outputs
  for a set of scalar inputs, cannot be used in this case.

\item The second, and related, challenge is the very large size of our
  data set. Though the number of simulations is small, each snapshot
  has over two million grid points as shown in
  Table~\ref{tab:sample_params}. In addition, as the simulations are
  run on multiple processors, each snapshot is split across multiple
  files.  Any algorithms used to build a high-quality surrogate must
  be modified to account for both these factors.

\end{itemize}

This report focuses on the first challenge of building accurate
spatio-temporal surrogate models given a limited number of simulations.
The second challenge of processing the extremely-high dimensional
snapshots generated in a distributed manner is discussed in the
companion report~\cite{kamath2023:stmclus}.

%
\section{Related work}
\label{sec:related}
%

Our approach to creating accurate, two-dimensional, spatio-temporal
surrogates builds on some early work in turbulence and pattern
recognition, specifically the characterization and recognition of
human face images.  The early work of Sirovich and
Kirby~\cite{sirovich1987:faces,kirby1990:faces} showed that a data
matrix, formed by an ensemble of face images, similar to our
snapshots, can be transformed using the Karhunen-Lo\`eve expansion
(similar to the principal component
analysis (PCA)~\cite{jolliffe2016:pcareview}) such that each face is written
as a linear combination of two-dimensional basis functions, they
called ``eigenpictures''.  A close approximation to a face is then
obtained by truncating the linear combination to use only a small
number of the initial, more important, basis
functions 
and the corresponding weights, thus creating a lower-dimensional
representation. 
Kirby and Sirovich also applied their ideas to
problems in fluid flow, including data from simulations, and
introduced the snapshot method and the concept of
eigenflows~\cite{sirovich1987:snapshot,kirby1990:eigenflows}.

Following this early work, Turk and
Pentland~\cite{turk1991:cvpr,turk1991:eigenfaces} showed that 
these ideas enabled face recognition as we could recognize a new image
as a specific face if its weights matched those of the specific face.
They referred to the basis functions as ``eigenfaces'' as they were
obtained using eigenanalysis of the data matrix. In later work,
non-linear alternatives were explored to obtain a better
representation for data that did not necessarily lie on a linear
manifold. The techniques included locally-linear decompositions and
neural-network-based auto-encoders,
not only for face images, but also speech data in the form of time
series and images of handwritten
digits~\cite{hinton1997:manifold,kambhatla1993:fast,kambhatla1993:fastnldr,kambhatla1997:localpca}.

In problems where the data represents output snapshots from
simulations run with different input parameters, at possibly different
time steps, an obvious next step was to build a predictive model
relating the simulation input parameters and time step to the weights
characterizing a snapshot. This would enable the prediction of results
at parameters not included in the original set of simulations. Such an
approach was taken in the early work of Ly and Tran~\cite{ly2001:pod},
who used proper orthogonal decomposition (similar to PCA) for the
decomposition, and spline interpolation for the predictive model. A
similar idea was explored by Higdon et al.~\cite{higdon2008:highdim},
who also used PCA, but predicted the weights using Gaussian process
models, and by Swischuk et al.~\cite{swischuk2019:rom}, who compared
different machine learning models for predicting the weights.

These ideas have become the subject of much recent research,
especially as compute-intensive simulations have become an
increasingly important part of design and engineering, requiring rapid
generation of results. In particular, in the field on non-intrusive
reduced-order modeling
(ROM)~\cite{mitry2014:surrogatethesis,rajaram2020:randrom}, many
options have been proposed, both for the decomposition into a
lower-dimensional representation and for the predictive model that
relates this representation to the simulation inputs. While the
dimension reduction is often obtained using PCA, which is a linear
method, non-linear approaches developed in data
mining~\cite{lee2007:nldrbook}, have also been used, despite their
greater complexity.  These include locally-linear PCA, kernel PCA, and
deep neural
networks~\cite{amsallem2012:nonlocal,xing2016:manifold,aversano2019:pcakriging,fukami2020:nonlinrom}.
For the predictive models, a range of interpolation techniques have
been used, including radial basis function regression, Gaussian
processes, and deep neural
nets~\cite{yu2019:adaptiverom,wu2020:poddl,berzins2021:rom}.

Several unique aspects of our problem make it impractical to directly apply these ideas:

\begin{itemize}

\item {\bf The very large size of each snapshot:} Many of the problems
  considered in the literature for building spatio-temporal surrogates
  have data at spatial points that number in the thousands, or tens of
  thousands, while a few have hundreds of thousands of grid points.
  For example, among the larger-sized data sets, one test problem in
  fluid flow considered by Rajaram et al.~\cite{rajaram2020:randrom}
  had 1047 snapshots, each with 41,796 grid points and the other had
  1001 snapshots, each with 450,000 grid points. Both problems were
  static (no time dependence), with a low-rank, reduced-dimensional
  space.  A problem with more complex geometry considered by
  B{\=e}rzin{\v s} et al.~\cite{berzins2021:rom} had a structured grid
  with 2459 nodes and two data sets, one with 100 simulations and the
  other with 400 simulations, where each simulation was run for 100
  time steps.  In addition, 50 simulations each were generated for
  validation and testing. In
  contrast,   each snapshot in our data set has over 2 million grid points, which
  makes it challenging to both manage and process the data, requiring
  the algorithms to be modified
  suitably.

\item {\bf The small number of simulations and time steps:}
  Spatio-temporal models are typically built using a large number of
  simulations, numbering in the hundreds or thousands, as noted
  earlier. We found one example of a compute-intensive
  simulation~\cite{yeh2018:cpod} where just thirty simulations were
  run, each for 1000 time steps. While our constraints on resources
  also limit us to a similar number of simulations, we run each
  simulation for a much smaller number of time steps, as indicated in
  Table~\ref{tab:sample_params}.

\item {\bf The variable size of the domain:} Most problems considered
  in spatio-temporal modeling are formulated on a fixed domain, with
  grid points at the same fixed locations at all time steps across all
  simulations. There are a few exceptions; the work by Yeh et
  al.~\cite{yeh2018:cpod} considers a problem where the input
  parameters control the geometry of the domain, while the problem in
  B{\=e}rzin{\v s} et al.~\cite{berzins2021:rom} had a moving grid
  with a fixed number of grid points. As we show in
  Section~\ref{sec:preprocessing}, substantial pre-processing of our
  data is required to bring all snapshots into a common grid before we
  can build the surrogate.

\end{itemize}

Our contributions in this report are three-fold: First, we address the
issues above and show how we process the large number of grid points
per snapshot on domains that vary with each simulation.  Second, we
propose ways to determine the parameters for the algorithms used in
building the surrogates. Finally, we investigate the accuracy of
surrogates we might expect when the number of simulations and the
number of time steps at which we run each simulation are both small.
We show how we can use the small number of simulations to identify the
range of input parameters and to create a data set with a sufficient
diversity of outputs. We also consider simple ways in which we can
improve the quality of the surrogate while keeping the number of
simulations small.

%
\section{Solution approach}
\label{sec:approach}
%

Our approach to building accurate spatio-temporal surrogates for
jet-HE interaction problem is composed of multiple steps which we
describe in detail in the following sections. We started by carefully
generating sample points in the three-dimensional input parameter
space (Section~\ref{sec:sampling}).  We then pre-processed the output
from each simulation so that all snapshots from all simulations had
values defined at the same set of grid points, with the plate
locations aligned at the initial time step, $t = 0$
(Section~\ref{sec:preprocessing}). This allowed us to represent the
data in the form of a matrix, whose 1604 columns represented all the
snapshots from the 45 simulations, and each row represented the value
of the variable at a specific $(x,y)$ grid point location. This data
in the form of the snapshot matrix was used to build the
spatio-temporal model (Sections~\ref{sec:pca} and~\ref{sec:local_pca})
and the accuracy of the surrogate evaluated on new simulations
(Section~\ref{sec:results}).

%
\subsection{Sampling the input parameter space}
\label{sec:sampling}
%

One of the challenges in our problem is that simulating the
interaction of the jet with the HE is compute intensive, limiting the
number of sample points at which we can run the simulations.
Therefore, the location of these sample points has to be selected
carefully, which is challenging as we do not know a priori the outcome
of running the simulation at any sample point or the range of input
parameters we should use in generating the samples.

Therefore, we generated the sample points incrementally in the
three-dimensional input space using a modified version of the best
candidate algorithm~\cite{mitchell91:sampling,kamath2022:sampling}
that selects samples randomly, but far apart from each other. We first
generated a small set of 16 samples using a range of [5.0, 20.0]cm for
HE-length, [0.125, 0.25]cm for the jet radius, and [0.6,
0.95]cm/$\mu$s for the jet tip velocity. We then excluded four sample
points with HE-length greater than 16.2cm, as this length was too
large for the jet to even reach the plate at late time.  Next,
restricting HE-length to be in the range [5.0, 16.2]cm, we ran 12 new
samples. Of the resulting 24 samples, we found that we could shrink
the range of HE-length further to [5.0, 15.0]cm and also exclude
samples in the lower right corner of the (HE-length - jet-tip-velocity)
plot where the jet tip velocity was too low for the plate to break. This
left us with data from 20 simulations, from an initial set of 28
simulations.

Having identified the range of values for generating simulations, we
added 25 new samples in this region, for a total of 45 samples. As the
best-candidate method is a progressive sampling algorithm, it allows
us to add samples incrementally, while preserving the random and
far-from-each-other property of the samples. 

\begin{figure}[htb]
\centering
\begin{tabular}{cc}
\includegraphics[trim = 0cm 0cm 0cm 0.0cm, clip = true,width=0.5\textwidth]{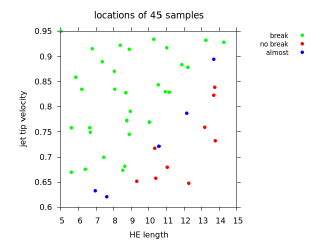} &
\includegraphics[trim = 0cm 0cm 0cm 0.0cm, clip = true,width=0.5\textwidth]{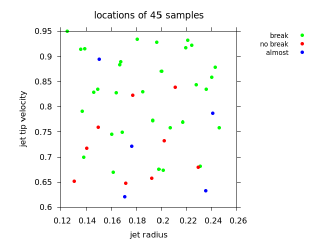} \\
\end{tabular}
\caption{The 45 samples in the space of the three input parameters for
  the simulation, labeled by the state of the plate at the last time
  step of the simulation. The simulation at the extreme corner of the
  input space, with HE-length, jet tip velocity, and jet radius equal to
  5.0cm, 0.950cm/$\mu$s, and 0.125cm, respectively, is referred to as
  the {\it baseline} simulation. It has the smallest number of time
  steps, with 29 snapshots.}
\label{fig:all_samples}
\end{figure}

Our data set, shown in Figure~\ref{fig:all_samples}, indicates that at
high jet tip velocity, but low HE-length, the plate breaks, while at low
jet tip velocity and high HE-length, the jet does not penetrate the plate.
This latter region is sparsely sampled as we are interested mainly in
cases where the plate breaks.
The class label (break, no break, or almost break) was assigned by
examining the outputs at the last two time steps in each simulation.

It is clear that our data set is unbalanced as we have 9 samples where
the plate does not break, 31 where the plate breaks, and 5 that are
almost break. 
Generating an appropriate data set for a problem like ours is
challenging as we can run only a limited number of simulations.
However, the boundary between the classes is poorly defined and we do
not know the range of inputs that will give us sample points with the
desired outcome. We erred on the side of having more break cases as
these were of greater interest. The outputs for the no-break cases
tended to be very similar, and we expected that a small number of such
cases would suffice.  Admittedly, as shown in
Section~\ref{sec:results}, our choice of sample points affects the
accuracy of the spatio-temporal surrogates built using the data set.

These 45 simulations generate a total of 1604 snapshots that vary in
the number of grid points as indicated for the four examples in
Table~\ref{tab:sample_params}. The simulation at the extreme corner of
the input space, with HE-length, jet tip velocity, and jet radius equal to
5.0cm, 0.950cm/$\mu$s, and 0.125cm, respectively, is referred to as
the {\it baseline} simulation. It has the smallest number of time
steps, with 29 snapshots.

%
\subsection{Pre-processing the data}
\label{sec:preprocessing}
%

To create a spatio-temporal surrogate for our data set, we first need
to create a {\it snapshot matrix} for each of the three output
variables, mass, x-momentum, and y-momentum. This snapshot matrix,
${\bf X} \in {\R}^{D \times N}$, is just a collection of the snapshots
\begin{equation}
{\bf X} = \left[ {\bf x}_1, {\bf x}_2, \ldots, {\bf x}_N \right] 
\label{eqn:smatrix}
\end{equation}
where ${\bf x}_i \in \R^D$, $N = 1604$ is the number of snapshots, and
$D$ is the number of grid points in a snapshot. However, a number of
issues have to be addressed before we can generate the snapshot matrix
where each snapshot has the same number of grid points at the same
$(x,y)$ coordinates. These issues, and our solution approach, are
discussed in detail in the companion report~\cite{kamath2023:stmclus} and
described briefly below.

\begin{figure}[htb]
\centering
\begin{tabular}{c}
\includegraphics[trim = 0.0cm 0cm 0cm 0.0cm, clip = true,width=0.9\textwidth]{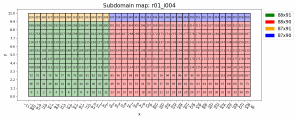} \\
\includegraphics[trim = 0.0cm 0cm 0cm 0.0cm, clip = true,width=0.9\textwidth]{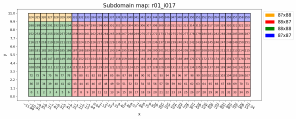} \\
\vspace{-0.6cm}
\end{tabular}
\caption{The distribution of data in 360 sub-domains for the first two
  example simulations in Table~\ref{tab:sample_params}. Top: key
  r01\_i004 and bottom: key r01\_i017. At each time step, the data in
  each sub-domain is available in a separate HDF5 file. 
  Note that the sizes of the sub-domains within a simulation vary and
  these sizes are different across simulations. All time steps in a
  simulation have the same distribution of the data.}
\label{fig:sample_subdomains}
\end{figure}

The data generated for each simulation, regardless of the size of the
domain, are available in 360 files in HDF5 format~\cite{hdf5:online}
for each time step, as shown in Figure~\ref{fig:sample_subdomains} for
two of the four example sub-domains.  Each HDF5 file includes five
variables --- $x$ and $y$ coordinates, mass, x-momentum, and
y-momentum. Within each file, the variables are in natural ordering,
that is, ordered by increasing values of the $y$-coordinate, and for a
fixed $y$-coordinate, ordered by increasing values of the
$x$-coordinate.  All simulations are on a regular grid with $\Delta x
= \Delta y = 0.0125$cm.

However, there are several differences in the data across the
simulations that preclude just appending the snapshots to create the
matrix ${\bf X}$. Figure~\ref{fig:sample_snaps_nmass} shows that each
simulation has a different domain size, with the same range of values
in $y$, but different ranges in $x$. In addition, while all
simulations write the output to 360 files, their sizes are different
as shown in Figure~\ref{fig:sample_subdomains}. We also observe that
the plate, which is a key structure in the problem, is at different
locations in the domain due to varying values of HE length in the
simulations. A closer look at the $(x,y)$ coordinates of the grid
points indicated that the values of $\Delta x$ and $\Delta y$ are not
exactly $0.0125$cm across simulations, and while the coordinates in
$y$, which has a fixed range of values, are identical across
simulations, this is not the case for the $x$ coordinates.

To address these differences across simulations, we first aligned the
domain for each simulation so that the origin was at the lower right
corner, which automatically aligned the plate at time $t=0$ across all
simulations. Next, we cropped the left end, removing data with the
shifted $x$ coordinate outside the range [-32, 0.0], which corresponds
to the smallest HE length of 5.0cm. Now, all snapshots have data on
the same domain, though not at the same $(x,y)$ coordinates.
To accomplish this, we re-mapped the data to a common grid, using a
simple 1-nearest neighbor algorithm. 

To make further processing of the large snapshot matrix feasible, we
generated it in a block form
\begin{equation}
{\bf X} =
\begin{bmatrix}
{\bf X}_{b1} \\
{\bf X}_{b2} \\
\ldots \\
{\bf X}_{bk} \\
\end{bmatrix}
\label{eqn:blockx}
\end{equation}
by using a block form for the common grid that was used to remap the
data. Each block was written to a separate file and contained all grid
points in a specific non-overlapping range of $y$ values, stored in
natural order.  Thus, concatenating the blocks in the order of their
$y$ values, would result in a single snapshot-matrix file for each
variable, with the grid points stored in natural order in the rows of
the matrix.

\begin{figure}[htb]
\centering
\setlength\tabcolsep{1pt}
\begin{tabular}{cc}
\includegraphics[width=0.45\textwidth]{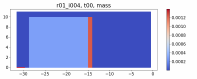} &
\includegraphics[width=0.45\textwidth]{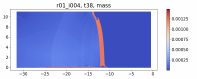} \\
\includegraphics[width=0.45\textwidth]{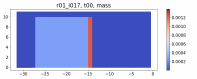} &
\includegraphics[width=0.45\textwidth]{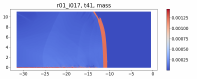} \\
\includegraphics[width=0.45\textwidth]{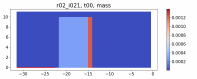} &
\includegraphics[width=0.45\textwidth]{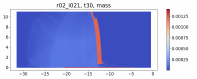} \\
\includegraphics[width=0.45\textwidth]{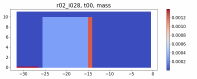} &
\includegraphics[width=0.45\textwidth]{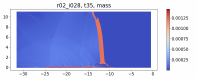} \\
\end{tabular}
\caption{The variable {\it mass}, after the snapshots have been
  aligned, cropped, and remapped. Left column shows the first time
  step and right column shows the last time step for the four example
  simulations in Table~\ref{tab:sample_params}. From top to bottom,
  simulations with keys r01\_i004, r01\_i017, r02\_i021, r02\_i028,
  respectively. The color bars are different between simulations and
  across time steps.}
\label{fig:sample_aligned_nmass}
\end{figure}

At the end of the pre-processing step, we have each of the three
output variables, in a separate snapshot matrix, where each matrix has
1604 columns and 2,180,799 rows. Each matrix is split by rows into 22
blocks, each block stored in a separate file.
Figure~\ref{fig:sample_aligned_nmass} shows the mass variable for the
first and last snapshot, for each of our four example simulations,
after the raw output data have been aligned, cropped, and remapped.
The corresponding images for x-momentum and y-momentum are shown in
Figures~\ref{fig:sample_aligned_nxmom}
and~\ref{fig:sample_aligned_nymom} in Appendix~\ref{sec:appendix1}
and~\ref{sec:appendix2}, respectively.

\begin{figure}[!htb]
\centering
\setlength\tabcolsep{1pt}
\begin{tabular}{cc}
\includegraphics[width=0.45\textwidth]{r01_i017_00000_t00_nmass.pdf} &
\includegraphics[width=0.45\textwidth]{r02_i028_00000_t00_nmass.pdf} \\
\includegraphics[width=0.45\textwidth]{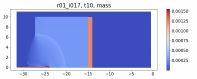} & 
\includegraphics[width=0.45\textwidth]{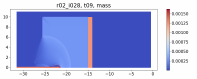} \\
\includegraphics[width=0.45\textwidth]{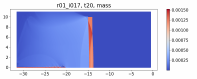} &
\includegraphics[width=0.45\textwidth]{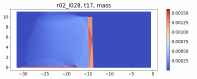} \\
\includegraphics[width=0.45\textwidth]{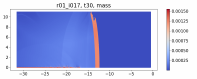} &
\includegraphics[width=0.45\textwidth]{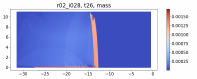} \\
\includegraphics[width=0.45\textwidth]{r01_i017_14157_t41_nmass.pdf} &
\includegraphics[width=0.45\textwidth]{r02_i028_13841_t35_nmass.pdf} \\
\end{tabular}
\caption{The variable {\it mass}, after the snapshots have been
  aligned, cropped, and remapped, at different time steps in two
  simulations showing the evolution of the data over time.  Left: key
  r01\_i017 (no break case) at time steps t0, t10, t20, t30, and t41.
  Right: key r02\_i028 (break case) at time steps t0, t09, t17, t26,
  and t35. The color bars are different between simulations and across
  time steps. }
\label{fig:timesteps_nmass}
\end{figure}

Next, to illustrate the evolution of the simulations over time, we
show the data, after the original output has been pre-processed, at
multiple time steps in two example simulations, key r01\_i017 and key
r02\_i028, corresponding to no-break and break cases, respectively.
Figure~\ref{fig:timesteps_nmass} shows the mass variable evolving
with time in these two simulations.  The corresponding images for
x-momentum and y-momentum are shown in
Figures~~\ref{fig:timesteps_nxmom} and ~\ref{fig:timesteps_nymom} in
Appendix~\ref{sec:appendix1} and~\ref{sec:appendix2}, respectively.

The pre-processing of the data described in this section is similar to
that performed in face recognition, where the face images are all
processed so they are the same size, with the face centered in the
image and major features such the eyes and nose aligned across
images. The difference in our data is that unlike a face, which is
stationary, the snapshots show the motion of the HE and the plate with
time. Therefore, even though we aligned the plate at time $t=0$ in all
simulations, there is a loss of alignment of the plate location as the
simulations progress, which will influence the accuracy of any
surrogate built from the data.

%
\subsection{Surrogate using a linear transformation}
\label{sec:pca}
%

After pre-processing the simulation output and converting it into a
snapshot matrix, ${\bf X}$, in a block form as shown in
Equation~\ref{eqn:blockx}, we build the spatio-temporal surrogate
using the traditional approach outlined in Section~\ref{sec:related}.
We start by linearly transforming the snapshot matrix ${\bf X}$ using
a singular value decomposition (SVD)~\cite{meyer2000:nabook} 
\begin{equation}
{\bf X} = {\bf U}  {\bf \Sigma}  {\bf V}^T 
\label{eqn:svd}
\end{equation}
where 
\begin{equation} \notag
{\bf X} \in \R^{D \times N}
\quad , \quad
{\bf U} \in \R^{D \times D}
\quad , \quad
{\bf \Sigma} \in \R^{D \times N}
\quad, \quad \text{and} \quad
{\bf V} \in \R^{N \times N} .
\end{equation}
Here, ${\bf \Sigma}$ is a diagonal matrix with non-zero diagonal
elements $\sigma_{ii}$, referred to as the singular values of the
matrix $\bf X$.  These singular values are typically ordered in
descending order, with the rows and the columns of the ${\bf U}$ and
${\bf V}$ matrices ordered correspondingly.  Since $D >> N$ in our
problem, the top $N$ rows of ${\bf \Sigma}$ will have non-zero
diagonal elements, assuming rank($\bf X$) = $N$. If the rank, $k$, is
less than $N$, then only the first $k$ diagonal elements will be
non-zero and the rest will be zero. The columns of the orthonormal
matrices ${\bf U}$ and ${\bf V}$ are referred to as the left and right
singular vectors of $\bf X$.  The matrix
\begin{equation}
{\bf U} = \left[ {\bf u}_1, {\bf u}_2, \ldots, {\bf u}_N \right] \quad \text{where} \quad {\bf u}_i \in \R^D
\end{equation}
is also the matrix of orthonormal eigenvectors of ${\bf X}{\bf X}^T$,
and the singular values $\sigma_i$ are the square-root of the
eigenvalues of ${\bf X}{\bf X}^T$ (or ${\bf X}^T{\bf X}$).  We refer
to the vectors ${\bf u}_i$ as ``{\it eigen-snapshots}'', akin to the
``eigenfaces'' used in face recognition~\cite{turk1991:eigenfaces}.

The ${\bf u}_i$ also form a basis in $\R^D$ for the data, so each snapshot ${\bf x}_i$
can be written as a linear combination of the ${\bf u}_k$ as follows
\begin{equation}
{\bf x}_i = \sum_{k=1}^N w_{ki} {\bf u}_k
\label{eqn:reconstruct_n}
\end{equation}
where the weight, $w_{ki}$, of the $k$-th basis for the $i$-th
snapshot is given as
\begin{equation}
w_{ki} = {\bf u}_k^T {\bf x}_i \quad \text{for} \quad k = 1, \ldots, N \ .
\label{eqn:wts}
\end{equation}
These weights are just the projection of the $i$-th
snapshot onto each of the columns of the $\bf U$ matrix.
Equation~\ref{eqn:reconstruct_n} shows how the original snapshot can
be reconstructed using the weights and the basis vectors.  A
reasonable approximation, $\tilde{\bf x}_i$, to a snapshot ${\bf x}_i$
can then be obtained by using the ${\bf u}_i$ corresponding to
the $n, n < N$, larger singular values:
\begin{equation}
\tilde{\bf x}_i = \sum_{k=1}^n w_{ki} {\bf u}_k .
\label{eqn:reconstruct_d}
\end{equation}

Thus, if we can generate the SVD for ${\bf X}$ as in
Equation~\ref{eqn:svd}, then Equations~\ref{eqn:wts}
and~\ref{eqn:reconstruct_d} enable us to generate an approximation,
$\tilde{\bf x}_i$, to the original snapshot, ${\bf x}_i$. These
equations can be interpreted as generating a reduced representation in
two ways - first, by creating an approximation that uses a smaller
number of weights and basis functions, ignoring the other weights and
basis functions, and second, by characterizing a snapshot ${\bf x}_i$
in terms of just its weights
\begin{equation}
{\bf w}_i = \left[ w_{1i},  w_{2i}, \ldots,  w_{Ni} \right] ,
\end{equation}
which can be combined with the basis set $( {\bf u}_k, k =
1, \ldots, N )$ to reconstruct the snapshots exactly.

This representation of the $i$-th snapshot, in terms of its weights,
$w_{ki}, k = 1, \ldots, N $, can be related to another representation
of the snapshot in terms of its simulation input
parameters $(h_i, v_i, r_i)$ and time step, $t_i$, through a predictive model
\begin{equation}
{w}_{ki} \approx f_k (h_i, v_i, r_i, t_i) , 
\text{ where }k = 1, \ldots, N \text{ and } i = 1, \ldots, N .
\end{equation}
Here $h_i, v_i$, and $ r_i$ are the HE length, jet tip velocity, and jet
radius for the $i$-th snapshot. Note that a separate model is created
for each weight index, that is, the first or most important weight is
predicted by one model, the second most important weight by another,
and so on, resulting in the $n$ (or $N$) models required to generate
the snapshots approximately (or exactly).  Then, to reconstruct
approximately the two-dimensional output, or snapshot at a new set of
input parameters and time step, we can first use the predictive model
for each of the $n$ weights to predict the weight values at this new
input and time step, and apply Equation~\ref{eqn:reconstruct_d}.

The spatio-temporal surrogate is thus composed of the basis functions
obtained using the SVD and the predictive models created using a
training set that relates the simulation inputs and time step to the
weights associated with the basis functions. There are several options
available to calculate the SVD and to generate the predictive models;
we next describe the options we selected for use in our problem.

We considered two ways to generate the SVD of ${\bf X}$ and obtain the
matrix ${\bf U}$ and the singular values $\sigma_i$:
\begin{itemize}

\item In the first approach based on the {\bf normal equations}, we
  form the matrix ${\bf X}^T {\bf X}$, of size $N \times N$, and
  obtain its eigen-decomposition. The singular value $\sigma_i$ of
  $\bf X$ is then the square-root of the eigenvalue $\lambda_i$ of
  ${\bf X}^T {\bf X}$. The columns ${\bf u}_i \in {\R}^{D}$ of $\bf U$
  are the corresponding orthonormal eigenvectors of ${\bf X} {\bf
    X}^T$. Since we have calculated the eigenvectors, ${\bf v}_i \in
  {\R}^{N} $, of the much smaller matrix ${\bf X}^T {\bf X}$, we can
  obtain the ${\bf u}_i $ as
  \begin{equation}
    {\bf u}_i = \frac{1}{\sqrt{\lambda_i}} {\bf X} {\bf v}_i \quad .
    \label{eqn:uafter}
  \end{equation}
  The division by $\sqrt{\lambda_i}$, which is the length of the
  vector ${\bf X} {\bf v}_i$, results in orthonormal vectors ${\bf
    u}_i$.

  This normal equations approach is very straight-forward and easily
  extends to the case when ${\bf X}$ is in block form
  (Equation~\ref{eqn:blockx}) as we can form the matrix ${\bf X}^T
  {\bf X}$ by reading in a block at a time. However, it does suffer
  from floating point issues associated with normal
  equations~\cite{meyer2000:nabook}.

\item A more stable approach to SVD is the {\bf QR
    decomposition} 
  that does not involve forming the normal equations.  Here, we first
  decompose the snapshot matrix
  as {\bf X} = {\bf Q}{\bf R}, where ${\bf Q} \in \R^{D \times D}$
  has orthonormal columns and ${\bf R}$ is upper triangular, and
  then generate the SVD of the smaller matrix $\bf R$. Since ${\bf X}$ is
  a tall, skinny matrix, we use the thin/reduced version of the QR
  decomposition~\cite{vandegeijn2023:ulaffbook}:

  \begin{equation}
    {\bf X} = {\bf Q}  {\bf R} =
    \begin{bmatrix}
      {\bf Q}_{1} & {\bf Q}_{2}
    \end{bmatrix}
    \begin{bmatrix}
      {\bf R}_{1} \\
      0 \\
    \end{bmatrix}
    = {\bf Q}_{1} {\bf R}_{1}
    \label{eqn:thinqr}
  \end{equation}
  where 
  \begin{equation} \notag
    {\bf X} \in \R^{D \times N}
    \quad , \quad
    {\bf Q} \in \R^{D \times D}
    \quad , \quad
    {\bf Q_1} \in \R^{D \times N}
    \quad , \quad
    {\bf R_1} \in \R^{N \times N} .
  \end{equation}
  We next obtain the SVD of the much smaller matrix ${\bf R_1}$ 
  \begin{equation}
    {\bf R}_1 = {\bf U}_{R_1}  {\bf \Sigma}_{R_1}  {\bf V}^T_{R_1} 
  \end{equation}
  where ${\bf U}_{R_1} \in \R^{N \times N}$, ${\bf \Sigma}_{R_1} \in
  \R^{N \times N}$, and ${\bf V}^T_{R_1} \in \R^{N \times N}$, which gives
  \begin{equation}
    {\bf X} = \bigl( {\bf Q}_{1} {\bf U}_{R_1} \bigr)  {\bf \Sigma}_{R_1}  {\bf V}^T_{R_1} .
  \end{equation}
  Thus, the singular values of ${\bf X}$ are just the singular values
  of ${\bf R}_1$ and the left singular vectors of ${\bf X}$. that is,
  the basis functions, are the columns of ${\bf Q}_1 {\bf U}_{R_1}$.

  To extend the QR decomposition to matrices in the block form, we
  used the work of Constantine and
  Gleich~\cite{constantine2011:blockqr}; this implementation is less
  straight-forward than the normal
  equations. 

\end{itemize}

We have several options for the $n$ predictive models to predict the
weights in the reduced representation of the data.  In our work, we
use a machine learning model, specifically, a Gaussian process
model~\cite{rasmussen06:book}, as it is one of the models that is
accurate for small data sets~\cite{kamath2018:regression}. It also
provides an estimate of the uncertainty on the prediction, which gave
us an opportunity to investigate whether the uncertainty could be used
to understand and explain the results, and to determine the number of
weights, $n$, to use in the reduced representation. The GP is an
expensive model to create, especially when we use automatic relevance
determination~\cite{rasmussen06:book}, where the hyper-parameters
include the weights on each input.  However, in our problem, we can
create and apply the $n$ models in parallel, which reduces the
turnaround time for creating the models.

%
\subsection{Surrogate using a locally-linear transformation}
\label{sec:local_pca}
%

The singular value decomposition, described in Section~\ref{sec:pca}
is a linear decomposition and if the data do not lie on a linear
manifold, the number of weights required for a reduced representation
may be quite large. This can create problems with accurate prediction
of the weights when we have a small number of simulations.

A simple approach to introducing nonlinearity in the decomposition is
to use a locally-linear decomposition, that is, cluster the snapshots
by similarity and then apply the linear, SVD-based approach to each
separate
cluster~\cite{kambhatla1993:fast,kambhatla1997:localpca,hinton1997:manifold}.
Then, to predict the two-dimensional output at a new point in the
simulation input space and time step, we identify which cluster the
new point belongs to, and use the predictive model created for that
cluster.

A challenge in clustering our data is the extreme high-dimensionality
of the snapshots as each snapshot has over 2.2M grid points. As
described in the companion report~\cite{kamath2023:stmclus}, we
investigated multiple ways to address this problem. For iterative
clustering methods, such as k-means, which require multiple passes
through the data, processing large files in each iteration can be
expensive. We considered two ways to avoid this processing: in the
first approach, we reduced the dimension prior to clustering by using
random projections~\cite{dasgupta2002:jlproof,li2006:randproj} and, in
the second approach, we clustered a different, but equivalent,
representation of the each snapshot, such as the weights obtained
after SVD (Equation~\ref{eqn:wts}). We also investigated a
non-iterative algorithm, hierarchical
clustering~\cite{jain88:clusbook}, which requires the pairwise
distances between the snapshots. By calculating the distance matrix
once, we could experiment with different options for the
method. We found that hierarchical clustering with Ward linkage gave
remarkably similar results to k-means clustering. We chose the results
with k-means and random projections as we could exploit the
randomness of both the random projections and the initial choice of
cluster centroids in k-means to identify the number of clusters.

A locally-linear transformation that combines clustering of snapshots,
following by an SVD on each cluster, is not the only option for
non-linearly transforming our data. Other options include kernel
PCA~\cite{xing2016:manifold}, which requires the solution of an
ill-posed inverse problem to reconstruct the two-dimensional output at
a new point in the simulation input space, and a neural-network based
auto-encoder, which, given the high-dimensionality of the snapshots,
will be a challenge to implement, even with a small number of hidden
layers~\cite{kambhatla1997:localpca}. In this work, we selected the
locally-linear transformation as it is a simple method and allowed the
re-use of software developed for the linear transformation.

%
\section{Experiments, results, and discussion}
\label{sec:results}
%

We next present the results of our experiments to test the accuracy of
our two spatio-temporal surrogates, one using a linear decomposition
and the other using a locally-linear decomposition.  Recall that the
approach in both cases is similar: we start with a snapshot matrix,
generate an SVD for it, and calculate the eigen-snapshots and the
corresponding weights that can be used to perfectly reconstruct each
snapshot.  Next, we build Gaussian process models that can predict
each of the initial, important weights based on the simulation inputs
and time step.  Then, to reconstruct an approximation to a new
snapshot that is identified by its simulation input parameters and
time step, we first predict the weights and then combine the weights with the
eigen-snapshots.  The two surrogates differ in the snapshot matrix
that is used - the linear surrogate uses a matrix composed of all
snapshots, while for the locally-linear surrogate, we cluster the
snapshots, grouping similar snapshots together, and then build a linear
surrogate for each cluster separately.

Our eventual goal in building these spatio-temporal surrogates is to
predict what happens to the jet and the plate at late time,
specifically, does the jet go through the plate, what is the velocity
of the jet tip when it goes through the plate, and what is the
location of the plate at late time. As the simulations are expensive,
we want to build accurate surrogates using only a small number of data
points in the simulation input space. As stated earlier, in this
initial study, we qualitatively evaluate the surrogates to understand
to what extent we can achieve our goals and to identify simple ways in
which we can improve the accuracy of the surrogates without
substantially increasing the number of simulations required.

In our work, given the questions we want to address regarding the
status of the plate and the jet at late time, we focus on the mass and
x-momentum variables. As shown in
Figures~\ref{fig:sample_aligned_nymom} and~\ref{fig:timesteps_nymom},
the y-momentum variable does not clearly define either the plate or
the jet, and is therefore less useful in our analysis.

In the following sections, we describe the test data we use to
evaluate the accuracy of reconstruction of new snapshots and discuss
how we determine various parameters used in our approach, including the
number of weights to use in the reconstruction, the number of clusters
to use for the locally-linear surrogate, and the identification the cluster
to which the new data point belongs. We also discuss ways to evaluate
the reconstruction quality for the test snapshots.

%
\subsection{Test simulations}
\label{sec:testdata}
%

\begin{figure}[tb]
\renewcommand{\arraystretch}{1.2}
  \begin{center}
    \begin{tabular}{|l|l|l|l|l|l|l|}
      \hline
      Simulation key & jet radius & HE-length & jet tip velocity & \texttt{\#} time &  outcome \\
                     & (cm)   & (cm)      & (cm/$\mu$sec) & steps  & \\
      \hline
      r03\_i008 & 0.221386  & 12.5439 & 0.725508 & 40 &  no break \\
      r03\_i018 & 0.153774  & 12.8638 & 0.830377 & 38 &  almost break \\
      r03\_i021 & 0.1647    & 11.7005 & 0.760524 & 38 &  almost break \\
      r03\_i023 & 0.197432  & 9.0411  & 0.613552 & 37 &  no break \\
      r03\_i026 & 0.219371  & 5.86578 & 0.912855 & 29 &  break \\ 
      r03\_i037 & 0.246562  & 9.47074 & 0.867685 & 33 &  break \\
      r03\_i050 & 0.211678  & 8.08681 & 0.794035 & 33 &  break \\
      \hline
    \end{tabular}
  \end{center}
  \captionof{table}{Input parameters for the seven test simulations
    used to evaluate the quality of the spatio-temporal surrogates.  }
  \label{tab:test_params}
\vspace{0.4cm}
\centering
\begin{tabular}{cc}
\includegraphics[trim = 0cm 0cm 0cm 0.0cm, clip = true,width=0.5\textwidth]{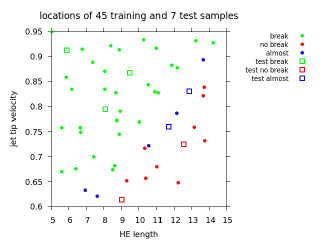} &
\includegraphics[trim = 0cm 0cm 0cm 0.0cm, clip = true,width=0.5\textwidth]{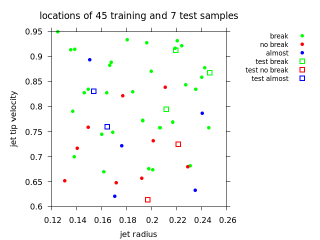} \\
\end{tabular}
\captionof{figure}{The 7 test simulations, along with the 45 training
  samples, in the space of the three input parameters, labeled by the
  state of the plate at the last time step of the simulation. Note
  that several of the test points are either near the boundary of the
  rectangular domain, or near the boundary of the region identified by
  the data points.  }
\label{fig:all_plus_test_samples}
\end{figure}

To test our ideas and evaluate the accuracy of the linear and
locally-linear surrogates created using the 45 simulations shown in
Figure~\ref{fig:all_samples}, we identified seven data points in this
simulation input parameter space. These points are listed in
Table~\ref{tab:test_params} and their locations are shown in
Figure~\ref{fig:all_plus_test_samples}.  The parameters for these data
points were selected to be different from each other, so we could
evaluate the quality of the predictions not only at different
locations in the input parameter space, but also at different
locations in the break, no-break, and almost-break space.  Three of
the seven points are clearly in the region where the plate breaks,
with two points each in the almost-break region and the no-break
regions. The latter four points were more difficult to identify as the
no-break region is less well defined in the (HE-length - jet-tip-velocity)
space, and the no-break points are near the boundary of the sampled
region. Their locations make it likely that the output predictions at
these four data points will be less accurate.

To enable us to evaluate the outputs predicted by the surrogates by
comparing with the actual outputs, we ran the simulation at these
seven points in the input space and identified the status of the plate
at the last time step. In practice, this last time step would be
identified as described in Section~\ref{sec:data_desc}.

%
\subsection{Generating the SVD}
\label{sec:svd}
%

There are several issues we need to address related to the
implementation of the SVD.  As explained in Section~\ref{sec:pca}, we
considered two implementations of the SVD algorithm - one based on the
normal equations and the other on a QR decomposition. In early
experiments, using a snapshot matrix with data from fewer than 45
simulations, we found that when select snapshots were reconstructed
using all the weights, the error was greater when we used the normal
equations formulation. This was expected behavior given the floating
point issues associated with the normal equations. Therefore, all
results in this report were generated using the QR decomposition.

Another issue that is the subject of much confusion is whether the
snapshot matrix should be ``centered'', that is, the mean snapshot
subtracted from each snapshot, before calculation of the SVD.  As
observed by B{\=e}rzin{\v s} et al.~\cite{berzins2021:rom}, there are adherents
on both sides of the issue.
Jolliffe and Cadima~\cite{cadima2009:pca,jolliffe2016:pcareview} have
discussed the topic at length, providing examples of where centering
may or may not be meaningful.  Some authors~\cite{alexandris2017:pca},
taking a pragmatic approach, have evaluated the results both ways, and
selected the option that best met their needs. For our data set, the
mean of the snapshot matrix with all 1604 snapshots is dominated by
the plate at early time; this is because in most snapshots, the plate, which is
aligned across all simulations at $t = 0$, has barely moved from the
initial location.  Therefore, when the mean snapshot is subtracted
from the snapshots at late time (the ones of most interest to us), the
initial position of the plate appears clearly. As a result, more
eigen-snapshots are required to account for, say, 90\% of the
variation in the data. In addition, we found that the substantial
visual change in the late time snapshots after centering made it
difficult to determine how many weights to use in the reconstruction.

%
\subsection{Clustering the snapshots}
\label{sec:clustering}
%

Clustering the 1604 snapshots for each of the three variables proved
to be challenging for several reasons. In Section~\ref{sec:local_pca},
we discussed how we addressed the issue of the extremely
high-dimensionality of each snapshot. We also wanted to understand if
there was an inherent clustering in the collection of snapshots.  A
careful analysis of the snapshots, as shown in
Figures~\ref{fig:sample_aligned_nmass} and~\ref{fig:timesteps_nmass}
for mass, and Figures~\ref{fig:sample_aligned_nxmom}
and~\ref{fig:timesteps_nxmom} for x-momentum, indicated that the first
and last snapshots in a simulation are quite different, suggesting
that multiple clusters exist in the data.  However, clustering the
snapshots would result in some neighboring snapshots, which are very
similar, assigned to two different clusters, which suggested that the
clusters are not well separated.  This made it difficult to identify
the number of clusters and to evaluate the results of any clustering
algorithms.
 
\begin{figure}[htb]
\centering
\begin{tabular}{cc}
\includegraphics[width=0.5\textwidth]{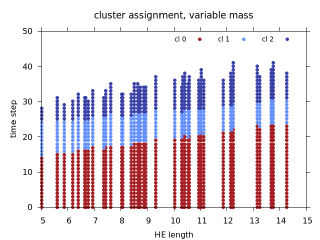} &
\includegraphics[width=0.5\textwidth]{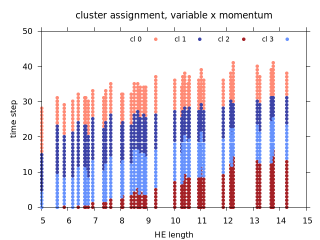} \\
\end{tabular}
\captionof{figure}{The cluster assignment, for the mass variable (3
  clusters) on the left, and the x-momentum variable (4 clusters) on
  the right. The plots show the times steps in the simulations plot
  along with the HE length. For the mass variable, the cluster sizes
  for clusters 0, 1, and 2, are 889, 351, and 364 snapshots,
  respectively.  For the x-momentum variable, the cluster sizes for
  clusters 0, 1, 2, and 3 are 397, 454, 305, and 448 snapshots,
  respectively. }
\label{fig:cass_mass_xmom}
\end{figure}

A detailed discussion on how we addressed these issues is given in the
companion report~\cite{kamath2023:stmclus}.
Figure~\ref{fig:cass_mass_xmom} shows the clustering results we use
for the mass and x-momentum variables. These results were obtained
using the k-means clustering algorithm, combined with random
projections to reduce the dimensionality of the snapshot matrix.
Based on these results, for our problem, it is relatively simple to
identify the cluster for each of the snapshots at which we want to
predict the 2-D output. Our interest is in the late time cluster,
which is cluster 2 for the mass variable and cluster 0 for the
x-momentum variable. Using these clusters, we can predict the output
at both the last time step and the time step that is 2 prior to the
last time step; 
we refer to these time steps as tlast and (tlast-2). Our reason for
predicting two late-time snapshots in each of the seven test
simulations will become clear in Section~\ref{sec:results1}.

%
\subsection{Determining the number of singular values to keep}
\label{sec:numsigma}
%

One of the key issues in reconstructing the simulation outputs at a
new data point is the determination of $n$, which is the number of
weights and basis vectors to use in building the approximation.
Ideally, we want a good approximation to a snapshot using a small
value for $n$. Let ${\bf X}_n$ be the matrix composed of snapshots
reconstructed using only the top $n$ weights.  For our problem, with a
tall, skinny $\bf X$, this error in the reconstruction is given by
\begin{equation}
\| {\bf X} - {\bf X}_n \|_F^2 = \sum_{i=n+1}^N \sigma_{ii}^2
\end{equation}

that is, the sum of squares of the singular values that were excluded
in the reconstruction. If these singular values are small, the error
in the approximation is small. The cumulative percentage variation
explained by the first $n$ singular vectors is
\begin{equation}
\frac{\sum_{ii=1}^n \sigma_{ii}^2}{\sum_{ii=1}^N \sigma_{ii}^2} * 100.0 \ .
\label{eqn:cumvar}
\end{equation}
Thus, one way to determine $n$ is to use a fixed percentage variation,
say 90\% or 95\%, that we would like explained by the reconstructed
data, and identify the $n$ associated with it. This is the most
popular approach in building spatio-temporal surrogates, where, having
identified $n$, predictive models are created for each of these $n$
weights and then used to predict the weights at a new data point.

\begin{table}[!htb]
  \begin{center}
    \begin{tabular}{|l|l|l|l|}
      \hline
      Data set & \# snapshots & 90\% & 95\% \\
      \hline
      mass, unclustered & 1604 & 6 & 17 \\
      mass, cluster 2 & 364 & 7 & 28 \\
      x-momentum, cluster 0 & 397 & 40 & 130 \\
      \hline
    \end{tabular}
  \end{center}
  \caption{Number of weights required to reconstruct the snapshot 
    matrix to account for 90\% and 95\% variation in the data for the mass variable (clustered and unclustered) and the x-momentum variable (clustered).  }
  \label{tab:num_weights}
\end{table}

\begin{figure}
\centering
\begin{tabular}{ccc}
\includegraphics[width=0.3\textwidth]{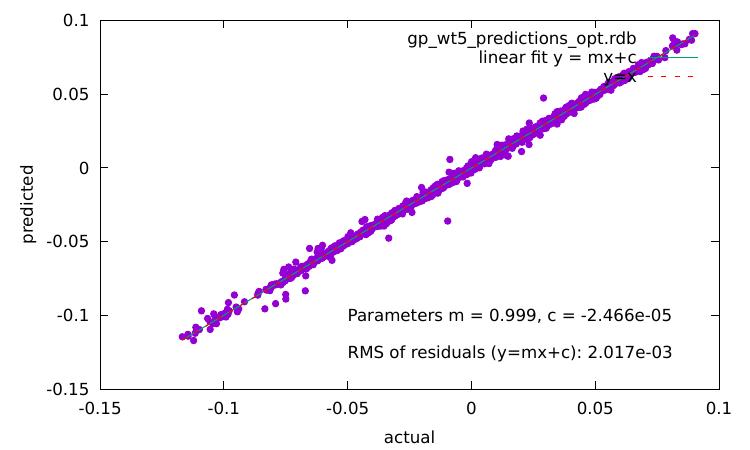} &
\includegraphics[width=0.3\textwidth]{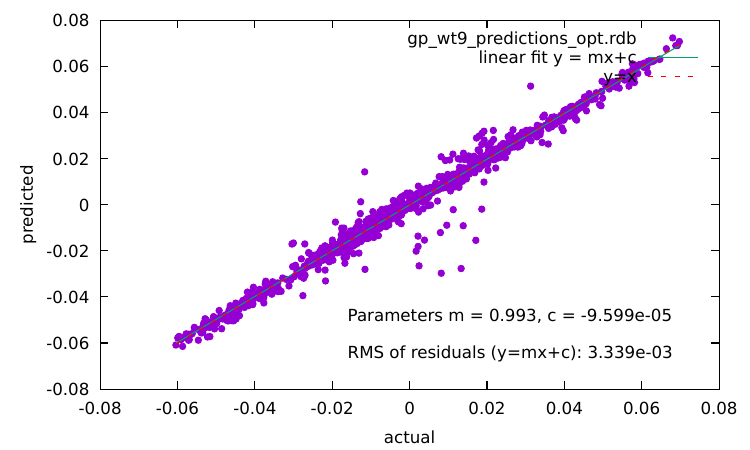} &
\includegraphics[width=0.3\textwidth]{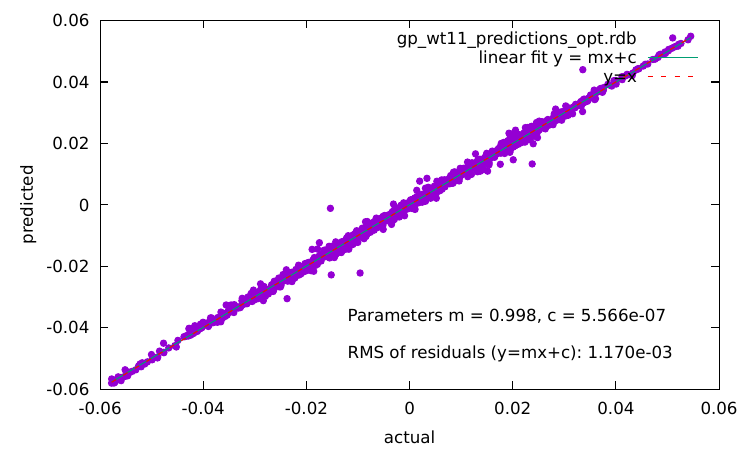} \\
(a) & (b) & (c) \\
\includegraphics[width=0.3\textwidth]{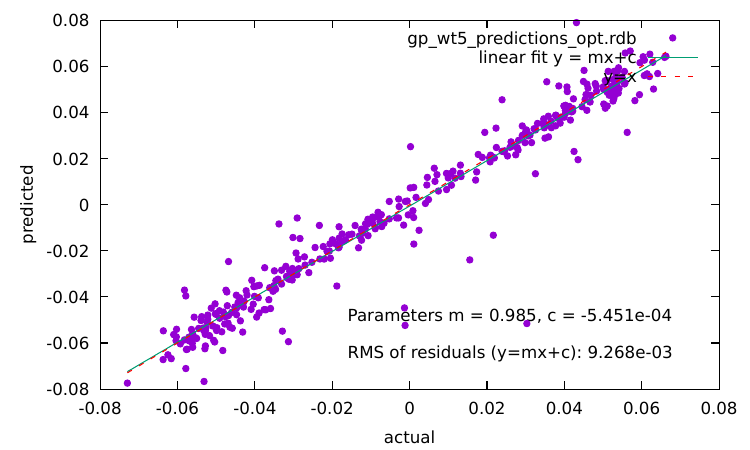} &
\includegraphics[width=0.3\textwidth]{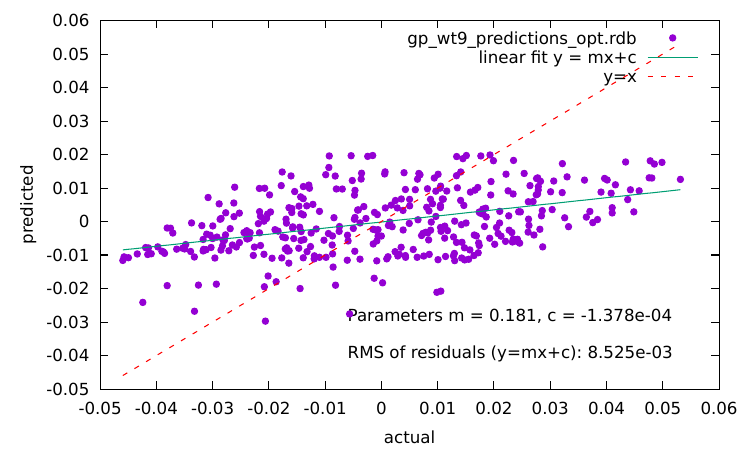} &
\includegraphics[width=0.3\textwidth]{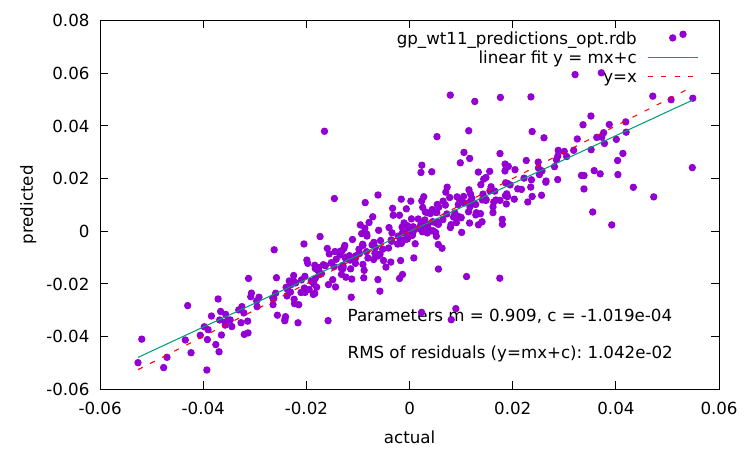} \\
(d) & (e) & (f) \\
\end{tabular}
\caption{Evaluating the prediction of the weights by the Gaussian
  process models using a plot of actual vs. predicted values obtained
  using leave-one-out validation.  Top row: mass, without clustering.
  Bottom: mass, cluster 2. From left to right, weight indices 5, 9,
  and 11 (the first weight has index 0). The red line is $y = x$ line,
  while the blue line is the best fit line to the data, whose slope
  and intercept is also included in each plot. The very early weights
  typically have little to no scatter as in (a), but the scatter
  increases with weight index, as in (b) and (d).  The prediction may
  be incorrect for most snapshots, as in (e). It may also improve with
  increasing weight index, as in (c) and (f).}
\label{fig:wts_act_pred}
\end{figure}

Table~\ref{tab:num_weights} lists the values of $n$ for our snapshot
matrices. As the values of $n$ can be quite large, and we have a small
number of simulations, we wanted to confirm the quality of the $n$
models with a leave-one-out approach prior to their use. We created
each model with all but one of the snapshots, and used the model to
predict the weight for the snapshot left out. For a good model, the
plot of the predicted vs.  actual weights should give points close to
the $y = x $ line.  However, we found that the quality of the models
tended to deteriorate quite quickly as the index of the weight
increased.  Figure~\ref{fig:wts_act_pred} shows a sample of these
predictions for the mass variable, with all 1604 snapshots and with
the 364 snapshots in the last cluster. We found that the most
important weights usually predicted very well, but the scatter around
the $y = x$ line increased with the weight index.  Sometimes, many of
the predicted values would be along the $y=x$ line, with some
relatively large outliers. In other cases, most of the predictions
were poor.  Interestingly, some weights at higher indices had better
predictions than the weights at lower indices.

We suspect that this poor prediction of some weights is due to the
small size of our data set, as we have just 45 simulations.
B{\=e}rzin{\v s} et al.~\cite{berzins2021:rom} observed that for one
of their problems, increasing the data set from 100 to 400
simulations did not affect the linear transformation, but
significantly improved the accuracy of the models created to predict
the weights.

The results in Figure~\ref{fig:wts_act_pred} indicated that we could
not use the quality of the actual vs. predicted weight for the training
data set to determine the number of weights for reconstruction. Even
when the overall quality of predictions on the training data are poor,
a weight prediction at a new data point could be accurate, and vice
versa.

We therefore decided to estimate the number of weights based on the
uncertainty in prediction obtained from the Gaussian process
surrogates. We predicted the values of the first 50 weights and
started by using the largest number of weights that all had small
variance in the predictions.  Since this number is relatively small
for the mass variable, we also generated the reconstructions using a
larger number of weights to understand the inaccuracies they would
introduce into the reconstructed outputs.

%
\subsection{Software and parallel implementation}
\label{sec:software}
%

Many of the steps in our approach to creating spatio-temporal
surrogates can be parallelized to reduce the computational time. In
this initial implementation, we exploited parallelism wherever it
was possible to do so easily. For example, tasks in the pre-processing
of the HDF5 files, such as reading the files, aligning and cropping
the domains, and remapping each simulation to the common grid, could
all be done in parallel across simulations using the Python sub-process
function. For the SVD using QR decomposition on our block snapshot
matrix, we implemented a serial version of the parallel algorithm by
Constantine and Gleich~\cite{constantine2011:blockqr}.
For the random projections used prior to clustering, we used a sparse
random matrix that was generated on the fly, reducing the storage
required, and enabling better use of the cache on the computer system.

Where possible, we used pre-existing, highly optimized software from
the double-precision BLAS~\cite{blackford2002:updatedblas} and
LAPACK~\cite{anderson1999:lapack} libraries. These included DGEMV and
DGEMM for matrix-vector and matrix-matrix multiply, respectively;
GEQRF, GESVD, and ORGQR for the SVD using the QR decomposition; and
DSYEVR for the eigenanalysis on the normal equations.

We also did not make any efforts to optimize the codes, for example by
finding the optimal block size to use in storing the snapshot matrix
or identifying ways in which consecutive steps in the processing could
be merged for faster turnaround.  Since our focus in this report is to
understand what is possible when the number of simulations is small,
we do not include compute times for various steps in our solution
approach.

%
\subsection{Reconstruction results}
\label{sec:results1}
%

We next present the results of reconstruction of the tlast and (tlast-2)
time steps for the seven test simulations using our two surrogates. For
the mass variable, we show results for both linear and locally-linear
transformations. As expected, the latter approach gave better results
as the models to predict the weights were created using snapshots that
were more similar to each other. Therefore, for the x-momentum
variable, we considered only the locally-linear transformations.

To determine the number of weights to use in reconstruction, we
generated the predictions and the uncertainties for the first 50
weights using the Gaussian process models. We expected, based on
Table~\ref{tab:num_weights}, that a maximum of 50 weights should
suffice to obtain a reasonable reconstruction for the mass and
x-momentum variables. Typically, the weights reduce in value with
increasing index number, and later weights that are near zero can be
dropped. However, it is possible that for a specific snapshot, a later
weight is too large to ignore.  We combined the information in
Table~\ref{tab:num_weights} with the uncertainties in the weight
predictions, to determine the exact number of weights to use in
reconstruction as follows:

\begin{itemize}

\item {\bf mass variable using the linear transformation:} Using the
  SVD results for all 1604 snapshots, we reconstructed the test
  snapshots with 8, 20, and 47 weights. 
  We considered the first 8 weights as 
  these weights were predicted with low uncertainty for all 14 test snapshots and
  Table~\ref{tab:num_weights} indicated that we would capture more
  than 90\% of the variation in the training data. We considered 20
  weights as 17 weights allowed us to account for atleast 95\%
  variation in the training data, and adding weights 18 through 20,
  which were predicted with low variation for all 14 test snapshots,
  could only improve the results.  Finally, we used 47 weights to
  understand the effects of adding more weights; we stopped at 47 as
  all test snapshots had higher uncertainty on prediction of weight
  48. We observe that for many test snapshots, the six weights
  with indices ranging from 9 to 15, often had high uncertainty. We
  therefore expected reconstructions with more than 8 weights to be of
  poor quality, but we wanted to understand to what extent including
  weight predictions with larger uncertainties would influence the
  results.

\item {\bf mass variable using the locally-linear transformation:}
  Using the SVD results for the 364 snapshots in the late-time cluster, we
  generated the reconstructed test snapshots with 30 weights. The errors in the
  initial weights tended to be small, especially for the (tlast-2)
  time steps. The first large error typically occurred at weight 31,
  resulting in our choice of 30 weights for reconstruction.

\item {\bf x-momentum variable using the locally-linear
    transformation:} Table~\ref{tab:num_weights} indicated we needed
  40 weights to capture 90\% variation in the reconstructed snapshots.
  We used all 50 weights as the error in the prediction of weights at
  higher indices was often very small.

\end{itemize}

We next describe the reconstruction results using the two
spatio-temporal surrogates. First, in
Appendix~\ref{sec:appendix_nmass_evec}, we show the first ten
eigen-snapshots for the mass variable generated with the linear and
locally-linear transformations.  The corresponding eigen-snapshots for
the x-momentum variable, locally-linear transformation, are shown in
Appendix~\ref{sec:appendix_nxmom_evec}. In addition, for each of the
14 snapshots being reconstructed, we present the following detailed
results:

\begin{itemize}

\item {\bf the predictions obtained from the Gaussian process
    surrogate:} We plot the predictions for the first 50 weights, with
  the uncertainty in prediction shown as an error bar at 1 standard
  deviation. 

\item {\bf the original snapshot, along with various reconstructed
    snapshots:} We focus on the region around the plate as we want to
  predict what happens to the plate at late time.

\item {\bf a $y$-lineout using the actual values in the reconstructed
    snapshots:} This plot of the variable values at $ y = 6.0063$ and
  $-15.5 \le x \le -10.5$ (in our transformed coordinates) for the
  original and reconstructed snapshots gives a concise summary of the
  quality of reconstruction and makes the comparison more quantitative
  than comparing entire snapshots visually.

\end{itemize}

\begin{figure}[!htb]
\centering
\begin{tabular}{cc}
\includegraphics[width=0.48\textwidth]{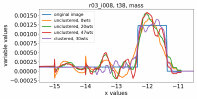} &
\includegraphics[width=0.48\textwidth]{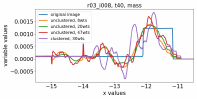} \\
\\
\includegraphics[width=0.48\textwidth]{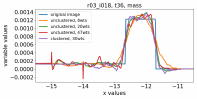} &
\includegraphics[width=0.48\textwidth]{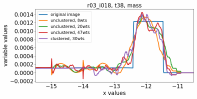} \\
\\
\includegraphics[width=0.48\textwidth]{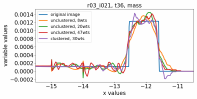} &
\includegraphics[width=0.48\textwidth]{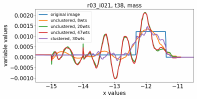} \\
\\
\includegraphics[width=0.48\textwidth]{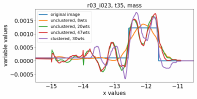} &
\includegraphics[width=0.48\textwidth]{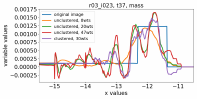} \\
\end{tabular}
\caption{The $y$-lineouts for comparing the reconstruction of the {\it
    mass} variable both with and without clustering. From top to
  bottom: test simulations {\bf r03\_i008}, {\bf r03\_i018}, {\bf
    r03\_i021}, and {\bf r03\_i023}, representing no break, almost
  break, almost break, and no break cases, respectively. Results for
  snapshot at time step (tlast-2) are shown on the left and time step tlast on
  the right.  }
\label{fig:recon_lineout_nmass_1}
\end{figure}

\begin{figure}
\centering
\begin{tabular}{cc}
\includegraphics[width=0.48\textwidth]{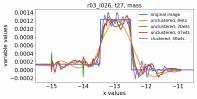} &
\includegraphics[width=0.48\textwidth]{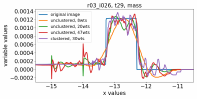} \\
\\
\includegraphics[width=0.48\textwidth]{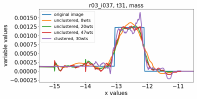} &
\includegraphics[width=0.48\textwidth]{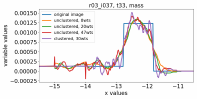} \\
\\
\includegraphics[width=0.48\textwidth]{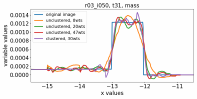} &
\includegraphics[width=0.48\textwidth]{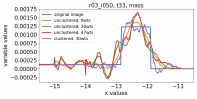} \\
\end{tabular}
\caption{The $y$-lineouts for comparing the reconstruction of the {\it
    mass} variable both with and without clustering. From top to
  bottom: test simulations {\bf r03\_i026}, {\bf r03\_i037}, and {\bf
    r03\_i050}, all representing break cases. Results for snapshot at
  time step (tlast-2) are shown on the left and time step tlast on the right.  }
\label{fig:recon_lineout_nmass_2}
\end{figure}

These detailed results, including weight predictions, reconstructed
snapshots, and the $y$-lineouts for the linear and locally-linear
surrogates for the mass variable at time steps tlast and (tlast-2) for
all seven test snapshots are shown in Appendix~\ref{sec:appendix3}.
The corresponding results for the x-momentum with the locally-linear
surrogate are shown in Appendix~\ref{sec:appendix4}. We repeat the
$y$-lineout for the reconstruction of the 14 test snapshots for the
mass variable in Figures~\ref{fig:recon_lineout_nmass_1}
and~\ref{fig:recon_lineout_nmass_2} and use them to discuss the
results and compare the different options used.


\clearpage

%
\subsection{Discussion}
\label{sec:discussion}
%

We next summarize our observations on the reconstructed results
presented for the mass and the x-momentum variables in
Figures~\ref{fig:recon_lineout_nmass_1}
and~\ref{fig:recon_lineout_nmass_2}, and in
Appendices~\ref{sec:appendix_nmass_evec} through~\ref{sec:appendix4}.
These results indicate the following:

\begin{itemize}

\item For the mass variable, the reconstructions of the plate region
  using the linear surrogate are usually worse than the locally-linear
  surrogate. This is observed visually in the reconstructed snapshots
  and more clearly in the $y$-lineouts.

  When we use all snapshots in the linear surrogate, the location of
  the plate at $t = 0$ is captured prominently in the early
  eigen-snapshots. However, all the test snapshots are at late time,
  when the plate has moved to the right from its original location.
  Therefore, any errors in the predicted values of the early weights,
  when multiplied by the corresponding eigen-snapshots, appear as
  vertical streaks in the region to the left of the plate in the
  reconstructed snapshots.  In contrast, for the locally-linear
  surrogate, the location of the plate in the eigen-snapshots is
  constrained to a smaller range of $x$ values and closer to where we
  might expect the plate to be at late time. This is to be expected as
  the snapshot matrix in the locally-linear surrogate includes only
  the late time snapshots, instead of all the snapshots in the linear
  surrogate.  The result is better quality reconstruction with the
  locally-linear surrogate.

  The creation of the locally-linear surrogate requires the additional
  step of clustering of the snapshots. However, the facts that the
  calculation of the SVD is for a smaller snapshot matrix and the
  Gaussian process models are built for a smaller data set, make the
  creation of the locally-linear surrogate computationally faster than
  the linear surrogate.

\item For both the mass and the x-momentum variables, the
  reconstruction at time step (tlast-2) is often better than at time
  step tlast.

  There are two contributing factors. First, the weights corresponding
  to higher indices at time step (tlast-2) are often smaller than the
  corresponding weights at time step tlast. So, when both snapshots
  are reconstructed with the same number of weights, ignoring the
  weights with higher indices, more useful information is ignored at
  time step tlast than at time step (tlast-2), resulting in better
  reconstruction of the latter.  Second, the error in prediction of
  the weights at time step (tlast-2) is usually smaller than at time
  step tlast. This is to be expected. When we consider the sample
  points in the input region formed by the three simulation parameters
  and the time step, the point at the last time step is on the
  boundary of this region, while the point at (tlast-2) time step is
  near, but not at the boundary. Weight predictions at the boundary
  points are usually less accurate as they have fewer neighboring
  points around them.

\item However, for the two no-break simulations, r03\_i008 and
  r03\_i023, the reconstructions of the mass and x-momentum variables,
  at both the tlast and the (tlast-2) time steps are poor.  

  These two simulations are near the boundary of the region formed by
  the training data in the space of the three input parameters, as
  shown in Figure~\ref{fig:all_plus_test_samples}.  Simulation
  r03\_i023 also has the smallest jet tip velocity. With fewer neighbors
  around these points, the weight predictions from the Gaussian
  process models are less accurate, resulting in poor quality reconstructions.

\item For the mass variable, we see different effects as we change the
  number of weights used in reconstruction with the linear surrogate.
  These effects are best understood through the $y$-lineplots. When
  the number of weights is small (=8), though the curve is very
  smooth, it is a poor fit to the plate profile, and extends far to
  the left of the plate. As the number of weights increases to 20 and
  then 47, the curves become a better fit to the profile of the plate
  and extend less to the left of the plate, but appear less smooth.

  There are several competing factors responsible for this behavior.
  The eigen-snapshots indicate that at a fixed $y$ coordinate the
  values flip between positive and negative, with fewer sign changes
  in the initial eigen-snapshots and more in the later ones, somewhat
  akin to a Fourier series. This accounts for the smooth curve in the
  $y$-lineout at few weights, which becomes more jagged as the number
  of weights is increased and the ``higher-frequency'' eigen-snapshots
  are used to approximate the plate, which can be seen as a square
  wave. As we increase the number of weights, the approximation to the
  plate location becomes better, and the $y$-lineout curve, which
  extends far to the left of the plate at 8 weights, moves closer to
  the plate at 20 and 47 weights. However, the weights between 8 and
  15 are often predicted with high error, which shows up as wiggles in
  the curve to the left of the plate. In contrast, for the
  locally-linear surrogate, the better localization of plate location
  in the eigen-snapshots and the lower error in weight prediction,
  lead to a better match of the reconstructed curve to the actual
  plate profile.

\item For the mass variable, the reconstructed snapshots have negative
  values, which appears physically incorrect. This is due to the
  limited number of eigen-snapshots used in reconstruction. Using all
  eigen-snapshots would result in near perfect reconstruction and all
  positive values (to within floating point errors) for the mass
  variable.

\item For the x-momentum variable, for which we only present results generated
  with the locally-linear surrogate, we observe certain differences
  with the mass variable. The values of both the variable and the
  weights are smaller for x-momentum. In most cases, the weights for
  x-momentum go rapidly to zero, though the initial weights have
  higher uncertainty. The reconstruction quality based on the
  $y$-lineplots indicates that the no-break cases and some of the
  break cases could be improved. 

\item We observe that in our problem, as the focus is on what happens
  to the plate and the jet at late time, we could have reduced the
  size of each snapshot further by focusing on just the region around
  the plate. However, such an option is problem dependent and may not
  be available in general.

\item Finally, we consider to what extent we can address the eventual
  goals of this effort, namely, for a new point in the simulation
  input space, is it possible to use the surrogate to determine if the
  plate breaks at late time, to identify the final location of the
  plate, and to obtain the speed of the jet as it exits the plate in
  cases where the plate breaks. 

  We address these questions using the reconstructed region around the
  plate and the $y$-lineouts for the cases where the snapshots have
  good reconstruction quality.  Understandably, the reconstructed
  snapshots are an approximation to the actual snapshots, especially
  when a small number of weights are used. However, the $y$-lineouts
  indicate that it should be possible to obtain a good estimate of the
  plate location at late time by applying gradient-based image
  segmentation techniques.  To determine whether the plate is a
  break, almost-break, or no-break case, we can look at the bottom
  region of the plate to see if it has detached from the bottom and if
  we can see the jet on the other side. We observe that it may be
  harder to distinguish between the no-break and the almost-break
  cases, but this may be due to the poor reconstruction of the
  no-break cases. However, the break cases appear to be easy to
  identify. This is despite the fact that the boundary of the plate is
  not as sharply defined in the reconstruction as we have ignored the
  weights at higher indices.

  Understanding whether we can meet our goals for the no-break cases,
  which are poorly reconstructed given our training data, we
  will first need to improve these cases using the ideas discussed
  next.

\end{itemize}

Overall, we observe that the quality of the reconstructed test
snapshots is a combination of several factors, including the
suitability of the initial eigen-snapshots at capturing the plate
location at late time, the error in the prediction of the weights, and
the number of weights used in the reconstruction. This gives us some
suggestions for generating better quality reconstructions, without
increasing the number of simulations:

\begin{itemize}

\item For our problem, the locally-linear surrogate created using only
  the late time snapshots, gives better results than the linear
  surrogate created using all snapshots from the 45 simulations.  We
  expect this result to hold in general. By clustering the snapshots,
  and building linear surrogates for each cluster, we generate a
  better basis for the snapshots in that cluster, leading to more
  accurate predictions for new snapshots in that cluster. This
  requires the identification of a cluster for the new snapshot, which
  can be obtained from the cluster assignment of all the snapshots. It
  also suggests that for our problem, other non-linear transform
  methods may be worth exploring.

\item To reduce the error in the prediction of the weights, we want
  the test point to be in the interior of the region formed by the
  training data in the space of the three input parameters and time
  step. 

  Therefore, if our interest is in predicting what happens at late
  time steps, as in our jet-HE interaction problem, we should run the
  simulations used to create the training data for a few more time
  steps, so the test points lie in the interior of the region that
  forms the input to the Gaussian process models.

  This also means that the test points should lie in the interior of
  the space of the simulation input parameters. Since the locations of
  the test points may not always be known before the training data are
  generated, one approach to ensure that the test points are not too
  far from the training data is to cover the region of interest with
  random points far from each other. We accomplish this using the best
  candidate sampling. In addition, decisions not to add points in some
  regions of the input space, should be taken with care; our decision
  to exclude points in the high HE-length, low jet tip velocity region led
  to poor prediction for the no-break test cases. Further, it may be
  desirable to set aside a number of simulations to be run once the
  locations of the test points are known.

\item It still remains a challenge to determine the number of weights
  to use for reconstruction, especially as the quality of the weight
  prediction could be poor when the number of simulations is small.
  Using too few weights would result in poor reconstruction of sharp
  changes in the data, such as the plate boundary, while using more
  weights might introduce errors when the weight predictions are not
  accurate.  Ideally, we want the weight values to decrease rapidly as
  it would indicate that a small number of initial weights is
  necessary for reconstruction.  However, this may not be the case
  when there is large variation in the data that is not captured
  sufficiently by a small number of simulations. It therefore appears
  that we may require experimentation with different number of
  weights, with the number selected possibly varying with each test
  snapshot.

\end{itemize}

%
\section{Conclusions}
\label{sec:conc}
%

In this report, we considered the problem of jet-HE interaction to
determine if it is possible to build accurate, spatio-temporal
surrogates when we can run only a small number of simulations to
create a training data set. We showed how to process a data set where
the size of the computational domain varies with each simulation and
each snapshot has over two million grid points. Our results showed
that a locally-linear surrogate, which builds separate surrogate
models using groups of similar snapshots, is more accurate than one
which builds a single surrogate using all the snapshots. We also
identified other simple ways to improve the quality of surrogates when
we are constrained to run only a limited number of simulations. These
include better sampling of the training data points in the simulation
input parameter space to cover the region uniformly so no test point
is too far from a training point; selecting, if possible, the
locations of the training data points such
that 
the test snapshots are not near the boundary of the region defined by
the training data; and setting aside a part of the simulation budget
to run a few additional simulations once the test data points have
been identified.

%
\section{Acknowledgment}
\label{sec:ack}
%

We would like to thank the Defense Threat Reduction Agency (DTRA) for
funding this work. The simulations of the interaction of the jet with
high explosives were performed using the ARES code developed at
Lawrence Livermore National Laboratory.

LLNL-TR-850152 This work performed under the auspices of the U.S. Department of
Energy by Lawrence Livermore National Laboratory under Contract
DE-AC52-07NA27344. 

This document was prepared as an account of work sponsored by an
agency of the United States government. Neither the United States
government nor Lawrence Livermore National Security, LLC, nor any of
their employees makes any warranty, expressed or implied, or assumes
any legal liability or responsibility for the accuracy, completeness,
or usefulness of any information, apparatus, product, or process
disclosed, or represents that its use would not infringe privately
owned rights. Reference herein to any specific commercial product,
process, or service by trade name, trademark, manufacturer, or
otherwise does not necessarily constitute or imply its endorsement,
recommendation, or favoring by the United States government or
Lawrence Livermore National Security, LLC. The views and opinions of
authors expressed herein do not necessarily state or reflect those of
the United States government or Lawrence Livermore National Security,
LLC, and shall not be used for advertising or product endorsement
purposes.

\bibliographystyle{plainnat}
\bibliography{ms_arxiv}

\clearpage

\appendix

%
\section{Appendix: Data for x-momentum variable}
\label{sec:appendix1}
%

\vspace{2cm}

\begin{figure}[htb]
\centering
\setlength\tabcolsep{1pt}
\begin{tabular}{cc}
\includegraphics[width=0.45\textwidth]{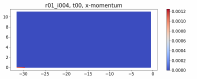} &
\includegraphics[width=0.45\textwidth]{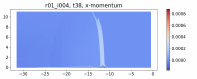} \\
\includegraphics[width=0.45\textwidth]{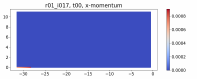} &
\includegraphics[width=0.45\textwidth]{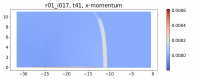} \\
\includegraphics[width=0.45\textwidth]{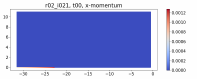} &
\includegraphics[width=0.45\textwidth]{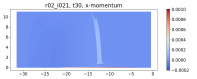} \\
\includegraphics[width=0.45\textwidth]{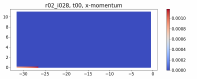} &
\includegraphics[width=0.45\textwidth]{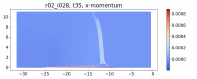} \\
\end{tabular}
\caption{The variable {\it x-momentum}, after the snapshots have been
  aligned, cropped, and remapped, at the first time step (left column)
  and last time step (right column) for the four example simulations
  in Table~\ref{tab:sample_params}. From top to bottom, simulations
  with keys r01\_i004, r01\_i017, r02\_i021, r02\_i028, respectively.
  The first time step has value zero for most of the domain, except
  the jet along the bottom on the left side of the region. The color
  bars are different between simulations and across time steps. }
\label{fig:sample_aligned_nxmom}
\end{figure}

\begin{figure}[htb]
\centering
\setlength\tabcolsep{1pt}
\begin{tabular}{cc}
\includegraphics[width=0.45\textwidth]{r01_i017_00000_t00_nxmom.pdf} &
\includegraphics[width=0.45\textwidth]{r02_i028_00000_t00_nxmom.pdf} \\
\includegraphics[width=0.45\textwidth]{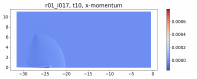} & 
\includegraphics[width=0.45\textwidth]{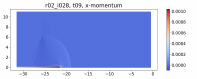} \\
\includegraphics[width=0.45\textwidth]{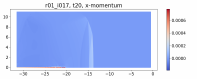} &
\includegraphics[width=0.45\textwidth]{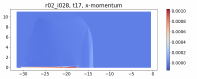} \\
\includegraphics[width=0.45\textwidth]{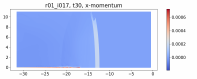} &
\includegraphics[width=0.45\textwidth]{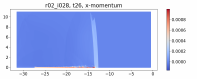} \\
\includegraphics[width=0.45\textwidth]{r01_i017_14157_t41_nxmom.pdf} &
\includegraphics[width=0.45\textwidth]{r02_i028_13841_t35_nxmom.pdf} \\
\end{tabular}
\caption{The variable {\it x-momentum}, after the snapshots have been
  aligned, cropped, and remapped, at different time steps in two
  simulations showing the evolution of the data over time.  Left: key
  r01\_i017 (no break case) at time steps t0, t10, t20, t30, and t41.
  Right: key r02\_i028 (break case) at time steps t0, t09, t17, t26,
  and t35. The color bars are different between simulations and across
  time steps.}
\label{fig:timesteps_nxmom}
\end{figure}

\clearpage

%
\section{Appendix:  Data for y-momentum variable}
\label{sec:appendix2}
%

\vspace{2cm}

\begin{figure}[htb]
\centering
\setlength\tabcolsep{1pt}
\begin{tabular}{cc}
\includegraphics[width=0.45\textwidth]{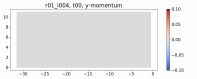} &
\includegraphics[width=0.45\textwidth]{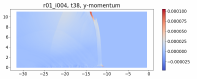} \\
\includegraphics[width=0.45\textwidth]{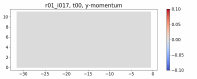} &
\includegraphics[width=0.45\textwidth]{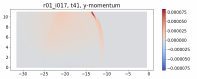} \\
\includegraphics[width=0.45\textwidth]{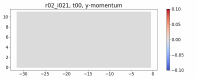} &
\includegraphics[width=0.45\textwidth]{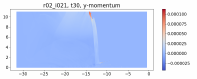} \\
\includegraphics[width=0.45\textwidth]{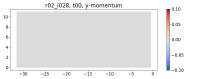} &
\includegraphics[width=0.45\textwidth]{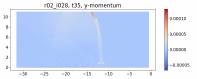} \\
\end{tabular}
\caption{The variable {\it y-momentum}, after the snapshots have been
  aligned, cropped, and remapped, at the first (left column) and last
  (right column) time steps for the four example simulations in
  Table~\ref{tab:sample_params}. From top to bottom, simulations with
  keys r01\_i004, r01\_i017, r02\_i021, r02\_i028, respectively. The
  first time step has value zero for most of the domain. Unlike the
  x-momemtum in Figure~\ref{fig:sample_aligned_nxmom}, there is no
  y-momentum for the jet at initial time. The color bars are different
  between simulations and across time steps.}
\label{fig:sample_aligned_nymom}
\end{figure}

\begin{figure}[htb]
\centering
\setlength\tabcolsep{1pt}
\begin{tabular}{cc}
\includegraphics[width=0.45\textwidth]{r01_i017_00000_t00_nymom.pdf} &
\includegraphics[width=0.45\textwidth]{r02_i028_00000_t00_nymom.pdf} \\
\includegraphics[width=0.45\textwidth]{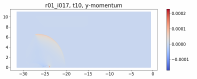} & 
\includegraphics[width=0.45\textwidth]{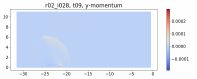} \\
\includegraphics[width=0.45\textwidth]{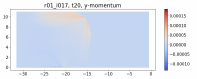} &
\includegraphics[width=0.45\textwidth]{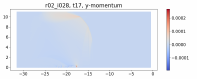} \\
\includegraphics[width=0.45\textwidth]{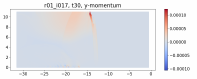} &
\includegraphics[width=0.45\textwidth]{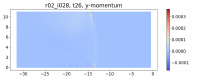} \\
\includegraphics[width=0.45\textwidth]{r01_i017_14157_t41_nymom.pdf} &
\includegraphics[width=0.45\textwidth]{r02_i028_13841_t35_nymom.pdf} \\
\end{tabular}
\caption{The variable {\it y-momentum}, after the snapshots have been
  aligned, cropped, and remapped, at different time steps in two
  simulations showing the evolution of the data over time.  Left: key
  r01\_i017 (no break case) at time steps t0, t10, t20, t30, and t41.
  Right: key r02\_i028 (break case) at time steps t0, t09, t17, t26,
  and t35. The color bars are different between simulations and across
  time steps.}
\label{fig:timesteps_nymom}
\end{figure}

\clearpage

%
\section{Appendix: Eigen-snapshots for mass variable, with and without clustering}
\label{sec:appendix_nmass_evec}
%

\begin{figure}[htb]
\centering
\setlength\tabcolsep{1pt}
\begin{tabular}{cc}
\includegraphics[width=0.45\textwidth]{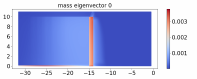} &
\includegraphics[width=0.45\textwidth]{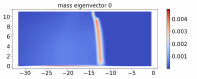} \\
\includegraphics[width=0.45\textwidth]{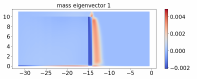} &
\includegraphics[width=0.45\textwidth]{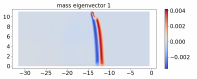} \\
\includegraphics[width=0.45\textwidth]{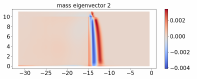} &
\includegraphics[width=0.45\textwidth]{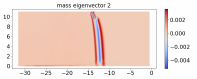} \\
\includegraphics[width=0.45\textwidth]{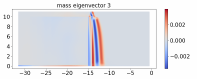} &
\includegraphics[width=0.45\textwidth]{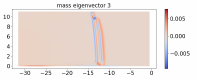} \\
\includegraphics[width=0.45\textwidth]{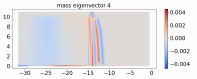} &
\includegraphics[width=0.45\textwidth]{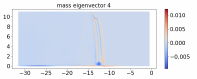} \\
\end{tabular}
\caption{Eigen-snapshots (0-4) for variable {\it mass}, without
  clustering (on left) and with clustering, cluster 2, on right. The
  color bars are different across eigen-snapshots.}
\label{fig:evec_nmass_a}
\end{figure}

\begin{figure}[htb]
\centering
\setlength\tabcolsep{1pt}
\begin{tabular}{cc}
\includegraphics[width=0.45\textwidth]{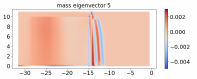} &
\includegraphics[width=0.45\textwidth]{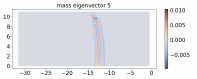} \\
\includegraphics[width=0.45\textwidth]{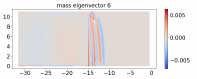} &
\includegraphics[width=0.45\textwidth]{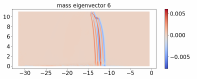} \\
\includegraphics[width=0.45\textwidth]{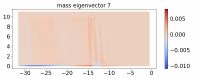} &
\includegraphics[width=0.45\textwidth]{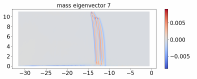} \\
\includegraphics[width=0.45\textwidth]{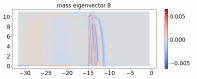} &
\includegraphics[width=0.45\textwidth]{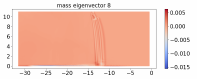} \\
\includegraphics[width=0.45\textwidth]{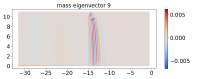} &
\includegraphics[width=0.45\textwidth]{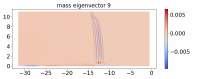} \\
\end{tabular}
\caption{Eigen-snapshots (5-9) for variable {\it mass}, without
  clustering (on left) and with clustering, cluster 2, on right. The
  color bars are different across eigen-snapshots.}
\label{fig:evec_nmass_b}
\end{figure}

\clearpage

%
\section{Appendix: Reconstructed mass (before and after clustering)
  for all seven test cases}
\label{sec:appendix3}
%


\begin{figure}
\centering
\begin{tabular}{cc}
\includegraphics[width=0.45\textwidth]{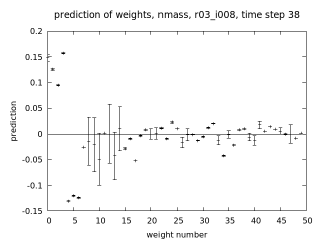} &
\includegraphics[width=0.45\textwidth]{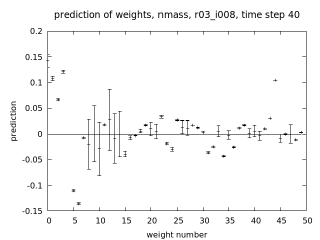} \\
\end{tabular}
\caption{Reconstruction of the variable {\it mass}, without
  clustering, for test simulation {\bf r03\_i008} at time t38 (left)
  and t40 (right). Plots show the weight predictions (with 1 standard
  deviation error bar) for the first 50 weights. }
\label{fig:recon_unclus_nmass_i008}
\vspace{0.5cm}
\centering
\begin{tabular}{cccc}
\includegraphics[width=0.20\textwidth]{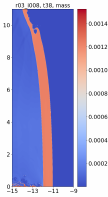} & 
\includegraphics[width=0.20\textwidth]{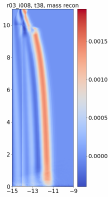} & 
\includegraphics[width=0.20\textwidth]{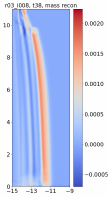} & 
\includegraphics[width=0.20\textwidth]{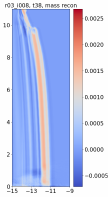} \\
\includegraphics[width=0.20\textwidth]{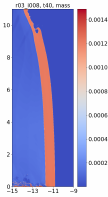} &
\includegraphics[width=0.20\textwidth]{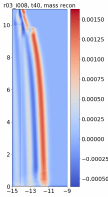} &
\includegraphics[width=0.20\textwidth]{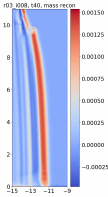} &
\includegraphics[width=0.20\textwidth]{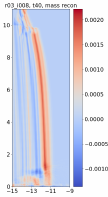} \\
\end{tabular}
\caption{ Reconstruction of the variable {\it mass}, without
  clustering, for test simulation {\bf r03\_i008} at time t38 (top)
  and t40 (bottom).  The plate view shows from left to right: the
  original snapshot followed by the reconstruction using 8, 20, and
  47 weights.  }
\label{fig:recon_unclus_nmass_i008_plate}
\end{figure}


\begin{figure}
\centering
\begin{tabular}{cc}
\includegraphics[width=0.45\textwidth]{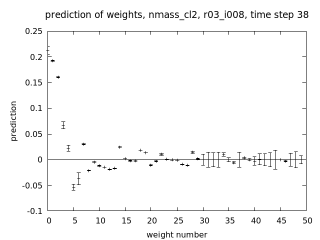} &
\includegraphics[width=0.45\textwidth]{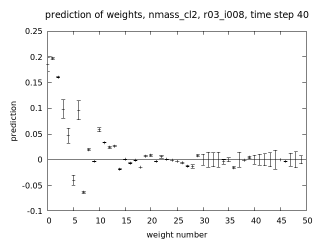} \\
\includegraphics[width=0.2\textwidth]{r03_i008_t38_nmass_plate_orig.pdf} 
\includegraphics[width=0.2\textwidth]{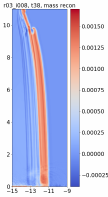} &
\includegraphics[width=0.2\textwidth]{r03_i008_t40_nmass_plate_orig.pdf} 
\includegraphics[width=0.2\textwidth]{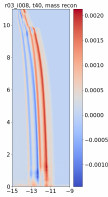} \\
\end{tabular}
\caption{Reconstruction of the variable {\it mass}, after clustering,
  for test simulation {\bf r03\_i008} at time t38 (left) and t40
  (right).  Top: the weight predictions (with 1 standard deviation
  error bar) for the first 50 weights. Bottom: each pair shows the
  plate view of the original snapshot and the reconstruction using 30
  weights.}
\label{fig:recon_clus_nmass_i008}
\end{figure}


\begin{figure}
\centering
\begin{tabular}{cc}
\includegraphics[width=0.48\textwidth]{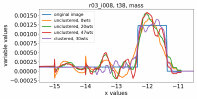} &
\includegraphics[width=0.48\textwidth]{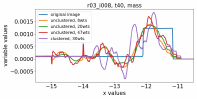} \\
\end{tabular}
\caption{Comparison of the reconstruction of the variable {\it mass},
  both with and without clustering, for test simulation {\bf
    r03\_i008} at time t38 (left) and t40 (right).  The plots compare the
  value of the variable at $y= 6.0063$ for the different
  reconstructions to indicate how well they detect the plate boundary. }
\label{fig:recon_lineout_nmass_i008}
\end{figure}



\begin{figure}
\centering
\begin{tabular}{cc}
\includegraphics[width=0.45\textwidth]{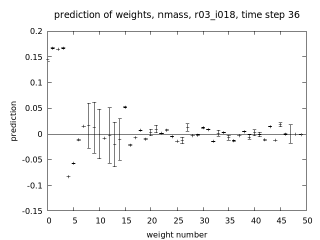} &
\includegraphics[width=0.45\textwidth]{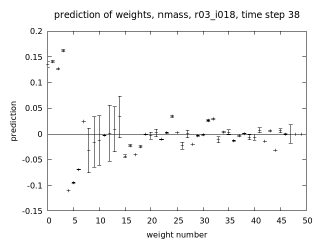} \\
\end{tabular}
\caption{Reconstruction of the variable {\it mass}, without
  clustering, for test simulation {\bf r03\_i018} at time t36 (left)
  and t38 (right). Plots show the weight predictions (with 1 standard
  deviation error bar) for the first 50 weights.}
\label{fig:recon_unclus_nmass_i018}
\vspace{0.5cm}
\centering
\begin{tabular}{cccc}
\includegraphics[width=0.2\textwidth]{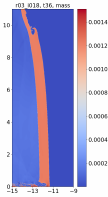} & 
\includegraphics[width=0.2\textwidth]{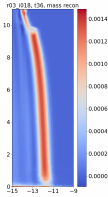} & 
\includegraphics[width=0.2\textwidth]{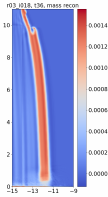} & 
\includegraphics[width=0.2\textwidth]{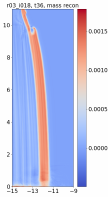} \\
\includegraphics[width=0.2\textwidth]{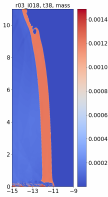} &
\includegraphics[width=0.2\textwidth]{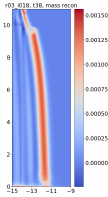} &
\includegraphics[width=0.2\textwidth]{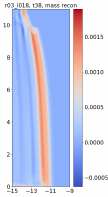} &
\includegraphics[width=0.2\textwidth]{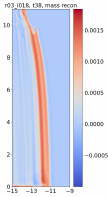} \\
\end{tabular}
\caption{ Reconstruction of the variable {\it mass}, without
  clustering, for test simulation {\bf r03\_i018} at time t36 (top)
  and t38 (bottom).  The plate view shows from left to right: the
  original snapshot followed by the reconstruction using 8, 20, and
  47 weights. }
\label{fig:recon_unclus_nmass_i018_plate}
\end{figure}


\begin{figure}
\centering
\begin{tabular}{cc}
\includegraphics[width=0.45\textwidth]{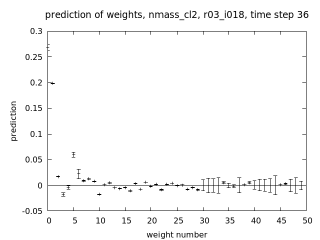} &
\includegraphics[width=0.45\textwidth]{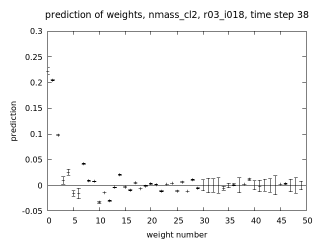} \\
\includegraphics[width=0.2\textwidth]{r03_i018_t36_nmass_plate_orig.pdf} 
\includegraphics[width=0.2\textwidth]{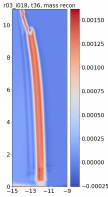} &
\includegraphics[width=0.2\textwidth]{r03_i018_t38_nmass_plate_orig.pdf} 
\includegraphics[width=0.2\textwidth]{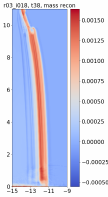} \\
\end{tabular}
\caption{Reconstruction of the variable {\it mass}, after clustering,
  for test simulation {\bf r03\_i018} at time t36 (left) and t38
  (right).  Top: the weight predictions (with 1 standard deviation
  error bar) for the first 50 weights. Bottom: each pair shows the
  plate view of the original snapshot and the reconstruction using 30
  weights. }
\label{fig:recon_clus_nmass_i018}
\end{figure}


\begin{figure}
\centering
\begin{tabular}{cc}
\includegraphics[width=0.48\textwidth]{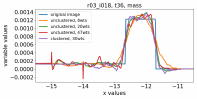} &
\includegraphics[width=0.48\textwidth]{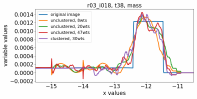} \\
\end{tabular}
\caption{Comparison of the reconstruction of the variable {\it mass},
  both with and without clustering, for test simulation {\bf
    r03\_i018} at time t36 (left) and t38 (right).  The plots compare the
  value of the variable at $y= 6.0063$ for the different
  reconstructions to indicate how well they detect the plate boundary.}
\label{fig:recon_lineout_nmass_i018}
\end{figure}



\begin{figure}
\centering
\begin{tabular}{cc}
\includegraphics[width=0.45\textwidth]{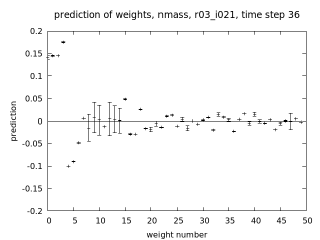} &
\includegraphics[width=0.45\textwidth]{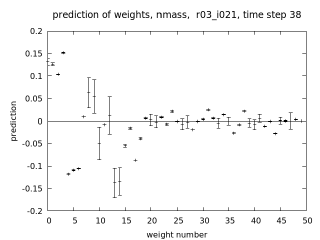} \\
\end{tabular}
\caption{Reconstruction of the variable {\it mass}, without
  clustering, for test simulation {\bf r03\_i021} at time t36 (left)
  and t38 (right).  Plots show the weight predictions (with 1 standard
  deviation error bar) for the first 50 weights. }
\label{fig:recon_unclus_nmass_i021}
\vspace{0.5cm}
\centering
\begin{tabular}{cccc}
\includegraphics[width=0.2\textwidth]{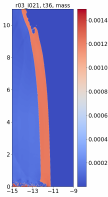} & 
\includegraphics[width=0.2\textwidth]{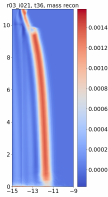} & 
\includegraphics[width=0.2\textwidth]{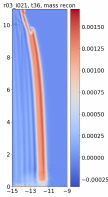} & 
\includegraphics[width=0.2\textwidth]{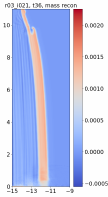} \\
\includegraphics[width=0.2\textwidth]{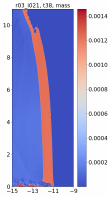} &
\includegraphics[width=0.2\textwidth]{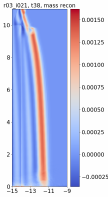} &
\includegraphics[width=0.2\textwidth]{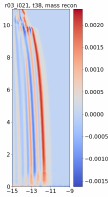} &
\includegraphics[width=0.2\textwidth]{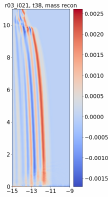} \\
\end{tabular}
\caption{Reconstruction of the variable {\it mass}, without
  clustering, for test simulation {\bf r03\_i021} at time t36 (top)
  and t38 (bottom).  The plate view shows from left to right: the
  original snapshot, followed by the reconstruction using 8, 20, and
  47 weights. }
\label{fig:recon_unclus_nmass_i021_plate}
\end{figure}


\begin{figure}
\centering
\begin{tabular}{ccc}
\includegraphics[width=0.45\textwidth]{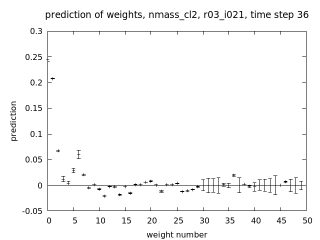} &
\includegraphics[width=0.45\textwidth]{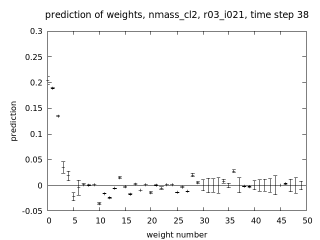} \\
\includegraphics[width=0.2\textwidth]{r03_i021_t36_nmass_plate_orig.pdf} 
\includegraphics[width=0.2\textwidth]{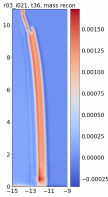} &
\includegraphics[width=0.2\textwidth]{r03_i021_t38_nmass_plate_orig.pdf} 
\includegraphics[width=0.2\textwidth]{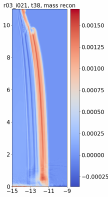} \\
\end{tabular}
\caption{Reconstruction of the variable {\it mass}, after clustering,
  for test simulation {\bf r03\_i021} at time t36 (left) and t38
  (right).  Top: the weight predictions (with 1 standard deviation
  error bar) for the first 50 weights. Bottom: each pair shows the
  plate view of the original snapshot and the reconstruction using 30
  weights. }
\label{fig:recon_clus_nmass_i021}
\end{figure}


\begin{figure}
\centering
\begin{tabular}{ccc}
\includegraphics[width=0.48\textwidth]{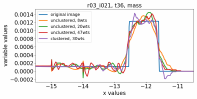} &
\includegraphics[width=0.48\textwidth]{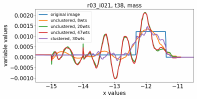} \\
\end{tabular}
\caption{Comparison of the reconstruction of the variable {\it mass},
  both with and without clustering, for test simulation {\bf
    r03\_i021} at time t36 (left) and t38 (right).  The plots compare the
  value of the variable at $y= 6.0063$ for the different
  reconstructions to indicate how well they detect the plate boundary.} 
\label{fig:recon_lineout_nmass_i021}
\end{figure}



\begin{figure}
\centering
\begin{tabular}{cc}
\includegraphics[width=0.45\textwidth]{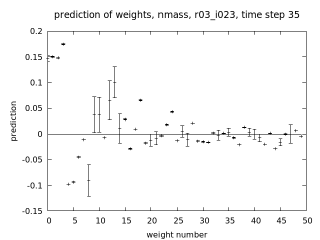} &
\includegraphics[width=0.45\textwidth]{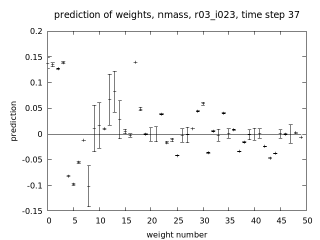} \\
\end{tabular}
\caption{Reconstruction of the variable {\it mass}, without
  clustering, for test simulation {\bf r03\_i023} at time t35 (left)
  and t37 (right).  Plots show the weight predictions (with 1 standard
  deviation error bar) for the first 50 weights. }
\label{fig:recon_unclus_nmass_i023}
\vspace{0.5cm}
\centering
\begin{tabular}{cccc}
\includegraphics[width=0.2\textwidth]{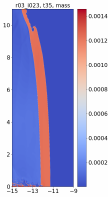} & 
\includegraphics[width=0.2\textwidth]{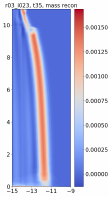} & 
\includegraphics[width=0.2\textwidth]{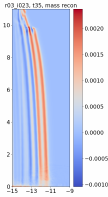} & 
\includegraphics[width=0.2\textwidth]{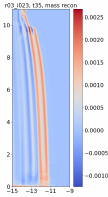} \\ 
\includegraphics[width=0.2\textwidth]{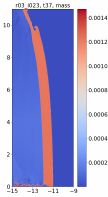} &
\includegraphics[width=0.2\textwidth]{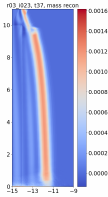} &
\includegraphics[width=0.2\textwidth]{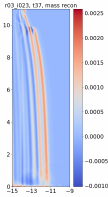} &
\includegraphics[width=0.2\textwidth]{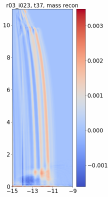} \\
\end{tabular}
\caption{Reconstruction of the variable {\it mass}, without
  clustering, for test simulation {\bf r03\_i023} at time t35 (top)
  and t37 (bottom).  The plate view shows from left to right: the
  original snapshot; followed by the reconstruction using 8, 20, and
  47 weights.}
\label{fig:recon_unclus_nmass_i023_plate}
\end{figure}


\begin{figure}
\centering
\begin{tabular}{ccc}
\includegraphics[width=0.45\textwidth]{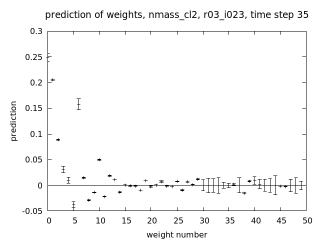} &
\includegraphics[width=0.45\textwidth]{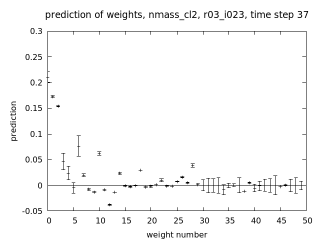} \\
\includegraphics[width=0.2\textwidth]{r03_i023_t35_nmass_plate_orig.pdf} 
\includegraphics[width=0.2\textwidth]{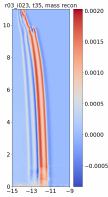} &
\includegraphics[width=0.2\textwidth]{r03_i023_t37_nmass_plate_orig.pdf} 
\includegraphics[width=0.2\textwidth]{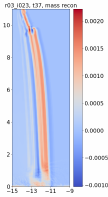} \\
\end{tabular}
\caption{Reconstruction of the variable {\it mass}, after clustering,
  for test simulation {\bf r03\_i023} at time t35 (left) and t37
  (right).  Top: the weight predictions (with 1 standard deviation
  error bar) for the first 50 weights. Bottom: each pair shows the
  plate view of the original snapshot and the reconstruction using 30
  weights. }
\label{fig:recon_clus_nmass_i023}
\end{figure}


\begin{figure}
\centering
\begin{tabular}{ccc}
\includegraphics[width=0.48\textwidth]{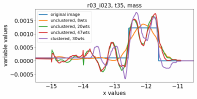} &
\includegraphics[width=0.48\textwidth]{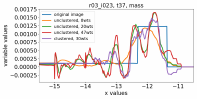} \\
\end{tabular}
\caption{Comparison of the reconstruction of the variable {\it mass},
  both with and without clustering, for test simulation {\bf
    r03\_i023} at time t35 (left) and t37 (right).  The plots compare the
  value of the variable at $y= 6.0063$ for the different
  reconstructions to indicate how well they detect the plate boundary.}
\label{fig:recon_lineout_nmass_i023}
\end{figure}



\begin{figure}
\centering
\begin{tabular}{cc}
\includegraphics[width=0.45\textwidth]{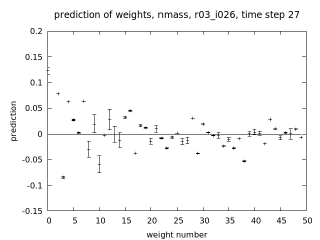} &
\includegraphics[width=0.45\textwidth]{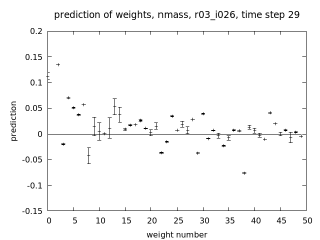} \\
\end{tabular}
\caption{Reconstruction of the variable {\it mass}, without
  clustering, for test simulation {\bf r03\_i026} at time t27 (left)
  and t29 (right).  Plots show the weight predictions (with 1 standard
  deviation error bar) for the first 50 weights. }
\label{fig:recon_unclus_nmass_i026}
\vspace{0.5cm}
\centering
\begin{tabular}{cccc}
\includegraphics[width=0.2\textwidth]{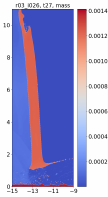} & 
\includegraphics[width=0.2\textwidth]{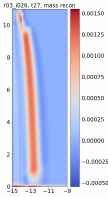} & 
\includegraphics[width=0.2\textwidth]{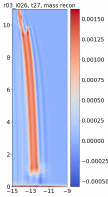} & 
\includegraphics[width=0.2\textwidth]{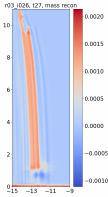} \\
\includegraphics[width=0.2\textwidth]{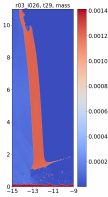} &
\includegraphics[width=0.2\textwidth]{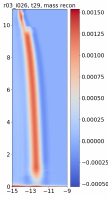} &
\includegraphics[width=0.2\textwidth]{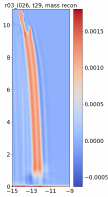} &
\includegraphics[width=0.2\textwidth]{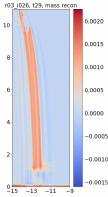} \\
\end{tabular}
\caption{Reconstruction of the variable {\it mass}, without
  clustering, for test simulation {\bf r03\_i026} at time t27 (top)
  and t29 (bottom).  The plate view shows from left to right: the
  original snapshot; followed by the reconstruction using 8, 20, and
  47 weights. }
\label{fig:recon_unclus_nmass_i026_plate}
\end{figure}


\begin{figure}
\centering
\begin{tabular}{cc}
\includegraphics[width=0.45\textwidth]{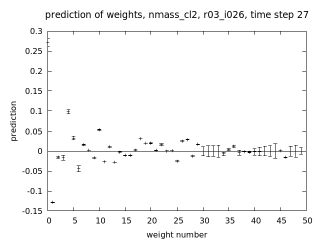} &
\includegraphics[width=0.45\textwidth]{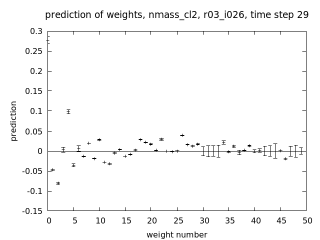} \\
\includegraphics[width=0.2\textwidth]{r03_i026_t27_nmass_plate_orig.pdf} 
\includegraphics[width=0.2\textwidth]{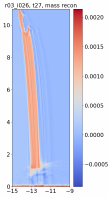} &
\includegraphics[width=0.2\textwidth]{r03_i026_t29_nmass_plate_orig.pdf} 
\includegraphics[width=0.2\textwidth]{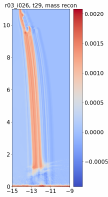} \\
\end{tabular}
\caption{Reconstruction of the variable {\it mass}, after clustering,
  for test simulation {\bf r03\_i026} at time t27 (left) and t29
  (right).  Top: the weight predictions (with 1 standard deviation
  error bar) for the first 50 weights. Bottom: each pair shows the
  plate view of the original snapshot and the reconstruction using 30
  weights. }
\label{fig:recon_clus_nmass_i026}
\end{figure}


\begin{figure}
\centering
\begin{tabular}{cc}
\includegraphics[width=0.48\textwidth]{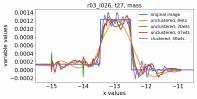} &
\includegraphics[width=0.48\textwidth]{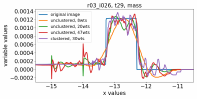} \\
\end{tabular}
\caption{Comparison of the reconstruction of the variable {\it mass},
  both with and without clustering, for test simulation {\bf
    r03\_i026} at time t27 (left) and t29 (right).  The plots compare the
  value of the variable at $y= 6.0063$ for the different
  reconstructions to indicate how well they detect the plate boundary.}
\label{fig:recon_lineout_nmass_i026}
\end{figure}



\begin{figure}
\centering
\begin{tabular}{cc}
\includegraphics[width=0.45\textwidth]{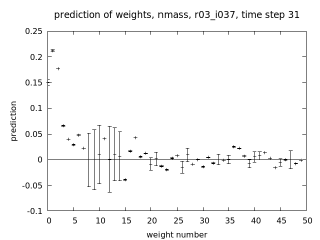} &
\includegraphics[width=0.45\textwidth]{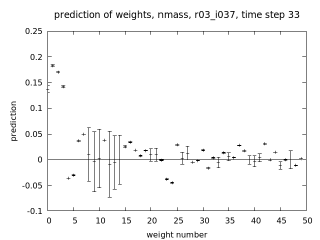} \\
\end{tabular}
\caption{Reconstruction of the variable {\it mass}, without
  clustering, for test simulation {\bf r03\_i037} at time t31 (left)
  and t33 (right).  The plots show the weight predictions (with 1
  standard deviation error bar) for the first 50 weights. }
\label{fig:recon_unclus_nmass_i037}
\vspace{0.5cm}
\centering
\begin{tabular}{cccc}
\includegraphics[width=0.2\textwidth]{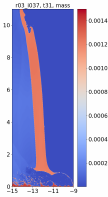} & 
\includegraphics[width=0.2\textwidth]{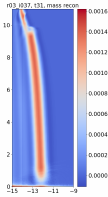} & 
\includegraphics[width=0.2\textwidth]{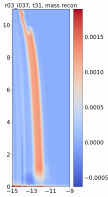} & 
\includegraphics[width=0.2\textwidth]{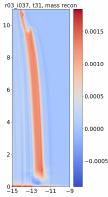} \\
\includegraphics[width=0.2\textwidth]{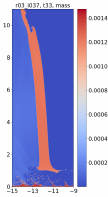} &
\includegraphics[width=0.2\textwidth]{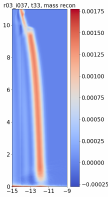} &
\includegraphics[width=0.2\textwidth]{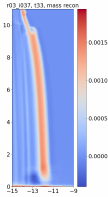} &
\includegraphics[width=0.2\textwidth]{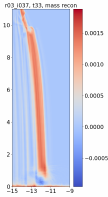} \\
\end{tabular}
\caption{Reconstruction of the variable {\it mass}, without
  clustering, for test simulation {\bf r03\_i037} at time t31 (top)
  and t33 (bottom).  The plate view shows from left to right: the
  original snapshot; followed by the reconstruction using 8, 20, and
  47 weights. }
\label{fig:recon_unclus_nmass_i037_plate}
\end{figure}


\begin{figure}
\centering
\begin{tabular}{cc}
\includegraphics[width=0.45\textwidth]{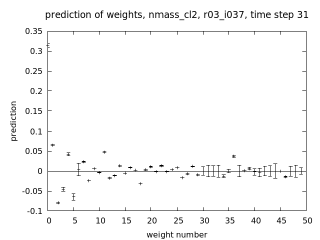} &
\includegraphics[width=0.45\textwidth]{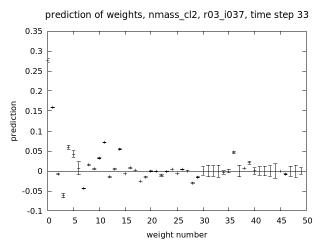} \\
\includegraphics[width=0.2\textwidth]{r03_i037_t31_nmass_plate_orig.pdf} 
\includegraphics[width=0.2\textwidth]{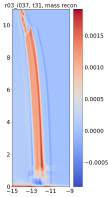} &
\includegraphics[width=0.2\textwidth]{r03_i037_t33_nmass_plate_orig.pdf} 
\includegraphics[width=0.2\textwidth]{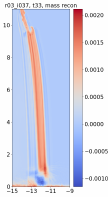} \\
\end{tabular}
\caption{Reconstruction of the variable {\it mass}, after clustering
  for test simulation {\bf r03\_i037} at time t31 (left) and t33
  (right). Top: the weight predictions (with 1 standard deviation
  error bar) for the first 50 weights. Bottom: each pair shows the
  plate view of the original snapshot and the reconstruction using 30
  weights. }
\label{fig:recon_clus_nmass_i037}
\end{figure}


\begin{figure}
\centering
\begin{tabular}{cc}
\includegraphics[width=0.48\textwidth]{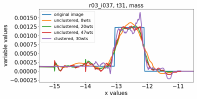} &
\includegraphics[width=0.48\textwidth]{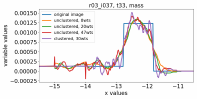} \\
\end{tabular}
\caption{Comparison of the reconstruction of the variable {\it mass},
  both with and without clustering, for test simulation {\bf
    r03\_i037} at time t31 (left) and t33 (right).  The plots compare the
  value of the variable at $y= 6.0063$ for the different
  reconstructions to indicate how well they detect the plate boundary.}
\label{fig:recon_lineout_nmass_i037}
\end{figure}



\begin{figure}
\centering
\begin{tabular}{cc}
\includegraphics[width=0.45\textwidth]{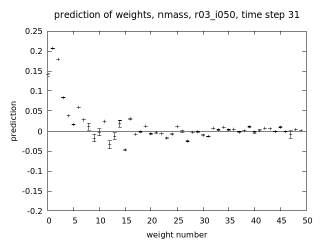} &
\includegraphics[width=0.45\textwidth]{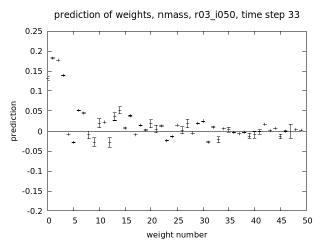} \\
\end{tabular}
\caption{Reconstruction of the variable {\it mass}, without
  clustering, for test simulation {\bf r03\_i050} at time t31 (left)
  and t33 (right).  Plots show the weight predictions (with 1 standard
  deviation error bar) for the first 50 weights. }
\label{fig:recon_unclus_nmass_i050}
\vspace{0.5cm}
\centering
\begin{tabular}{cccc}
\includegraphics[width=0.2\textwidth]{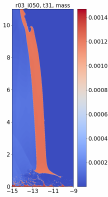} & 
\includegraphics[width=0.2\textwidth]{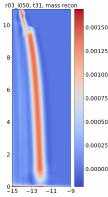} & 
\includegraphics[width=0.2\textwidth]{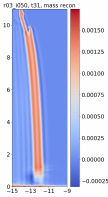} & 
\includegraphics[width=0.2\textwidth]{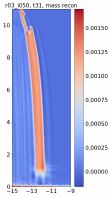} \\
\includegraphics[width=0.2\textwidth]{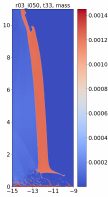} &
\includegraphics[width=0.2\textwidth]{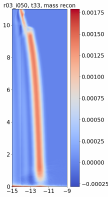} &
\includegraphics[width=0.2\textwidth]{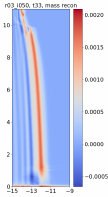} &
\includegraphics[width=0.2\textwidth]{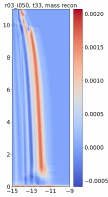} \\
\end{tabular}
\caption{Reconstruction of the variable {\it mass}, without
  clustering, for test simulation {\bf r03\_i050} at time t31 (top)
  and t33 (bottom).  The plate view shows from left to right: the
  original snapshot; followed by the reconstruction using 8, 20, and
  47 weights. }
\label{fig:recon_unclus_nmass_i050_plate}
\end{figure}


\begin{figure}
\centering
\begin{tabular}{cc}
\includegraphics[width=0.45\textwidth]{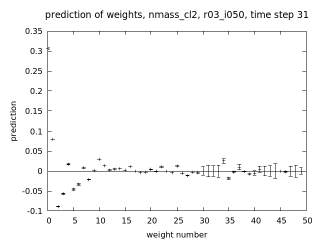} &
\includegraphics[width=0.45\textwidth]{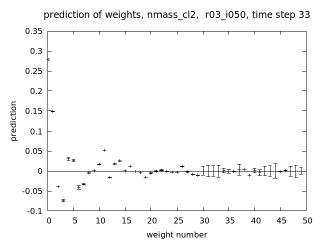} \\
\includegraphics[width=0.2\textwidth]{r03_i050_t31_nmass_plate_orig.pdf} 
\includegraphics[width=0.2\textwidth]{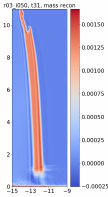} &
\includegraphics[width=0.2\textwidth]{r03_i050_t33_nmass_plate_orig.pdf} 
\includegraphics[width=0.2\textwidth]{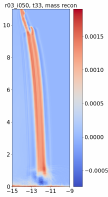} \\
\end{tabular}
\caption{Reconstruction of the variable {\it mass}, after clustering,
  for test simulation {\bf r03\_i050} at time t31 (left) and t33
  (right).  Top: the weight predictions (with 1 standard deviation
  error bar) for the first 50 weights. Bottom: each pair shows the
  plate view of the original snapshot and the reconstruction using 30
  weights. }
\label{fig:recon_clus_nmass_i050}
\end{figure}


\begin{figure}
\centering
\begin{tabular}{cc}
\includegraphics[width=0.48\textwidth]{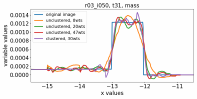} &
\includegraphics[width=0.48\textwidth]{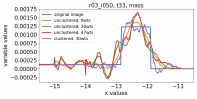} \\
\end{tabular}
\caption{Comparison of the reconstruction of the variable {\it mass},
  both with and without clustering, for test simulation {\bf
    r03\_i050} at time t31 (left) and t33 (right).  The plots compare the
  value of the variable at $y= 6.0063$ for the different
  reconstructions to indicate how well they detect the plate boundary.}
\label{fig:recon_lineout_nmass_i050}
\end{figure}

\clearpage

%
\section{Appendix: Eigen-snapshots for x-momentum variable, with clustering}
\label{sec:appendix_nxmom_evec}
%

\vspace{2cm}

\begin{figure}[htb]
\centering
\setlength\tabcolsep{1pt}
\begin{tabular}{cc}
\includegraphics[width=0.45\textwidth]{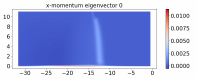} &
\includegraphics[width=0.45\textwidth]{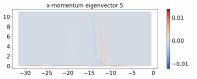} \\
\includegraphics[width=0.45\textwidth]{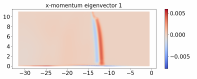} &
\includegraphics[width=0.45\textwidth]{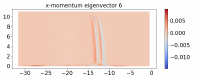} \\
\includegraphics[width=0.45\textwidth]{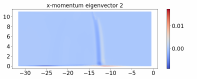} &
\includegraphics[width=0.45\textwidth]{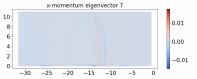} \\
\includegraphics[width=0.45\textwidth]{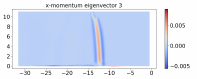} &
\includegraphics[width=0.45\textwidth]{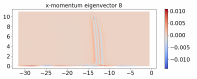} \\
\includegraphics[width=0.45\textwidth]{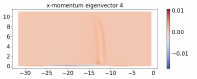} &
\includegraphics[width=0.45\textwidth]{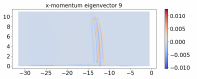} \\
\end{tabular}
\caption{Eigen-snapshots (0-4 on left and 5-9 on right) for variable
  {\it x-momentum}, with clustering. The color bars are different
  across eigen-snapshots.}
\label{fig:evec_nxmom}
\end{figure}

\clearpage

%
\section{Appendix: Reconstructed x-momentum (after clustering)
  for all seven test cases}
\label{sec:appendix4}
%

\clearpage


\begin{figure}
\centering
\begin{tabular}{ccc}
\includegraphics[width=0.45\textwidth]{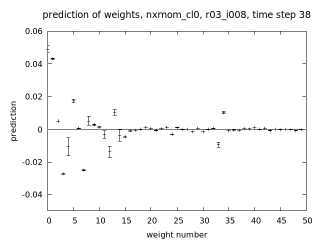} &
\includegraphics[width=0.45\textwidth]{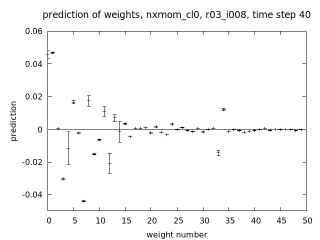} \\
\includegraphics[width=0.22\textwidth]{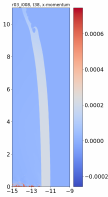} 
\includegraphics[width=0.22\textwidth]{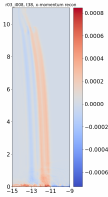} &
\includegraphics[width=0.22\textwidth]{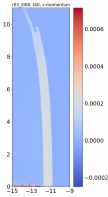} 
\includegraphics[width=0.22\textwidth]{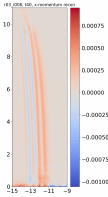} \\
\end{tabular}
\caption{Reconstruction of the variable {\it x-momentum}, after
  clustering, for test simulation {\bf r03\_i008} at time t38 (left)
  and t40 (right).  Top: the weight predictions (with 1 standard
  deviation error bar) for the first 50 weights. Bottom: each pair
  shows the plate view of the original snapshot and the reconstruction
  using 50 weights. The color bars are different across images.}
\label{fig:recon_clus_nxmom_i008}
\end{figure}


\begin{figure}
\centering
\begin{tabular}{cc}
\includegraphics[width=0.48\textwidth]{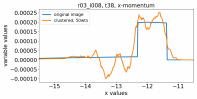} &
\includegraphics[width=0.48\textwidth]{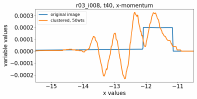} \\
\end{tabular}
\caption{Comparison of the reconstruction of the variable {\it
    x-momentum}, with clustering, for test simulation {\bf r03\_i008}
  at time t38 (left) and t40 (right).  The plots compare the value of
  the variable at $y= 6.0063$ for the reconstruction to
  indicate how well it detects the plate boundary.}
\label{fig:recon_lineout_nxmom_i008}
\end{figure}


\begin{figure}
\centering
\begin{tabular}{ccc}
\includegraphics[width=0.45\textwidth]{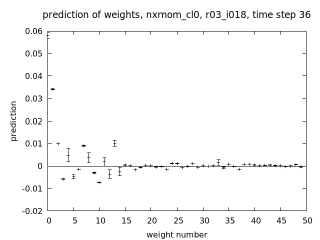} &
\includegraphics[width=0.45\textwidth]{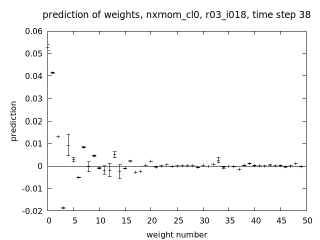} \\
\includegraphics[width=0.22\textwidth]{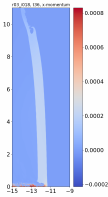} 
\includegraphics[width=0.22\textwidth]{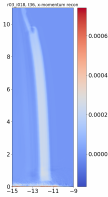} &
\includegraphics[width=0.22\textwidth]{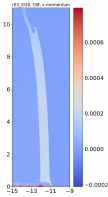} 
\includegraphics[width=0.22\textwidth]{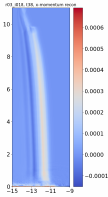} \\
\end{tabular}
\caption{Reconstruction of the variable {\it x-momentum}, after
  clustering, for test simulation {\bf r03\_i018} at time t36 (left)
  and t38 (right).  Top: the weight predictions (with 1 standard
  deviation error bar) for the first 50 weights. Bottom: each pair
  shows the plate view of the original snapshot and the reconstruction
  using 50 weights.  The color bars are different across images.}
\label{fig:recon_clus_nxmom_i018}
\end{figure}


\begin{figure}
\centering
\begin{tabular}{cc}
\includegraphics[width=0.48\textwidth]{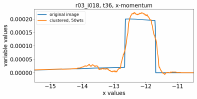} &
\includegraphics[width=0.48\textwidth]{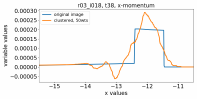} \\
\end{tabular}
\caption{Comparison of the reconstruction of the variable {\it
    x-momentum}, with clustering, for test simulation {\bf r03\_i018}
  at time t36 (left) and t38 (right).  The plots compare the value of
  the variable at $y= 6.0063$ for the reconstruction to
  indicate how well it detects the plate boundary.}
\label{fig:recon_lineout_nxmom_i018}
\end{figure}


\begin{figure}
\centering
\begin{tabular}{ccc}
\includegraphics[width=0.45\textwidth]{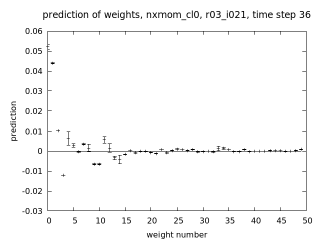} &
\includegraphics[width=0.45\textwidth]{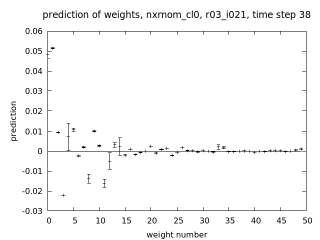} \\
\includegraphics[width=0.22\textwidth]{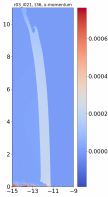} 
\includegraphics[width=0.22\textwidth]{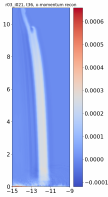} &
\includegraphics[width=0.22\textwidth]{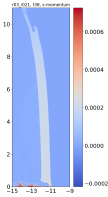} 
\includegraphics[width=0.22\textwidth]{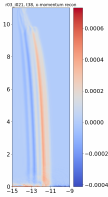} \\
\end{tabular}
\caption{Reconstruction of the variable {\it x-momentum}, after
  clustering, for test simulation {\bf r03\_i021} at time t36 (left)
  and t38 (right).  Top: the weight predictions (with 1 standard
  deviation error bar) for the first 50 weights. Bottom: each pair
  shows the plate view of the original snapshot and the reconstruction
  using 50 weights.  The color bars are different across images.}
\label{fig:recon_clus_nxmom_i021}
\end{figure}


\begin{figure}
\centering
\begin{tabular}{cc}
\includegraphics[width=0.48\textwidth]{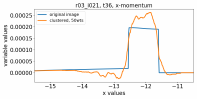} &
\includegraphics[width=0.48\textwidth]{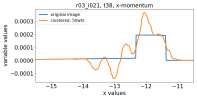} \\
\end{tabular}
\caption{Comparison of the reconstruction of the variable {\it
    x-momentum}, with clustering, for test simulation {\bf r03\_i021}
  at time t36 (left) and t38 (right).  The plots compare the value of
  the variable at $y= 6.0063$ for the reconstruction to indicate how
  well it detects the plate boundary.}
\label{fig:recon_lineout_nxmom_i021}
\end{figure}


\begin{figure}
\centering
\begin{tabular}{ccc}
\includegraphics[width=0.45\textwidth]{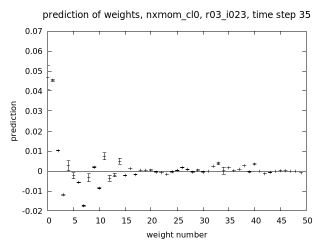} &
\includegraphics[width=0.45\textwidth]{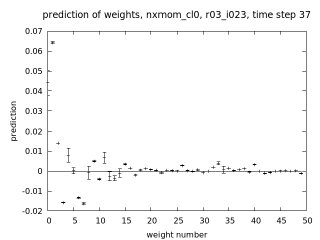} \\
\includegraphics[width=0.22\textwidth]{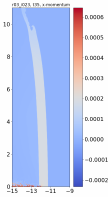} 
\includegraphics[width=0.22\textwidth]{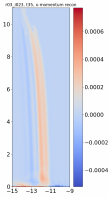} &
\includegraphics[width=0.22\textwidth]{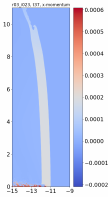} 
\includegraphics[width=0.22\textwidth]{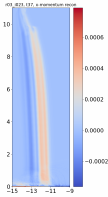} \\
\end{tabular}
\caption{Reconstruction of the variable {\it x-momentum}, after
  clustering, for test simulation {\bf r03\_i023} at time t35 (left)
  and t37 (right).  Top: the weight predictions (with 1 standard
  deviation error bar) for the first 50 weights. Bottom: each pair
  shows the plate view of the original snapshot and the reconstruction
  using 50 weights.  The color bars are different across images.}
\label{fig:recon_clus_nxmom_i023}
\end{figure}


\begin{figure}
\centering
\begin{tabular}{cc}
\includegraphics[width=0.48\textwidth]{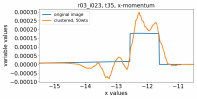} &
\includegraphics[width=0.48\textwidth]{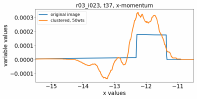} \\
\end{tabular}
\caption{Comparison of the reconstruction of the variable {\it
    x-momentum}, with clustering, for test simulation {\bf r03\_i023}
  at time t35 (left) and t37 (right).  The plots compare the value of
  the variable at $y= 6.0063$ for the reconstruction to indicate how
  well it detects the plate boundary.}
\label{fig:recon_lineout_nxmom_i023}
\end{figure}


\begin{figure}
\centering
\begin{tabular}{ccc}
\includegraphics[width=0.45\textwidth]{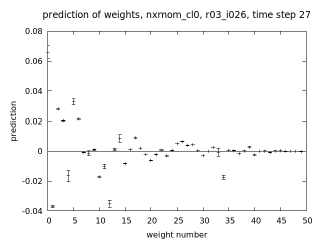} &
\includegraphics[width=0.45\textwidth]{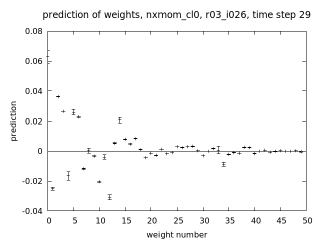} \\
\includegraphics[width=0.22\textwidth]{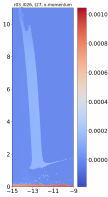} 
\includegraphics[width=0.22\textwidth]{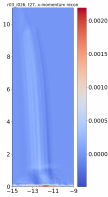} &
\includegraphics[width=0.22\textwidth]{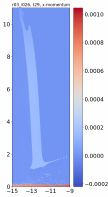} 
\includegraphics[width=0.22\textwidth]{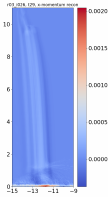} \\
\end{tabular}
\caption{Reconstruction of the variable {\it x-momentum}, after
  clustering, for test simulation {\bf r03\_i026} at time t27 (left)
  and t29 (right).  Top: the weight predictions (with 1 standard
  deviation error bar) for the first 50 weights. Bottom: each pair
  shows the plate view of the original snapshot and the reconstruction
  using 50 weights.  The color bars are different across images.}
\label{fig:recon_clus_nxmom_i026}
\end{figure}


\begin{figure}
\centering
\begin{tabular}{cc}
\includegraphics[width=0.48\textwidth]{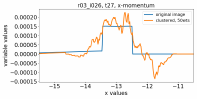} &
\includegraphics[width=0.48\textwidth]{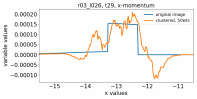} \\
\end{tabular}
\caption{Comparison of the reconstruction of the variable {\it
    x-momentum}, with clustering, for test simulation {\bf r03\_i026}
  at time t27 (left) and t29 (right).  The plots compare the value of
  the variable at $y= 6.0063$ for the reconstruction to
  indicate how well it detects the plate boundary.}
\label{fig:recon_lineout_nxmom_i026}
\end{figure}


\begin{figure}
\centering
\begin{tabular}{ccc}
\includegraphics[width=0.45\textwidth]{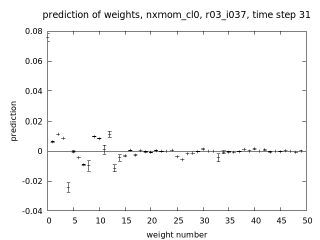} &
\includegraphics[width=0.45\textwidth]{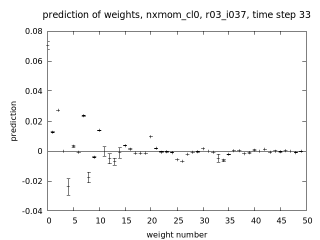} \\
\includegraphics[width=0.22\textwidth]{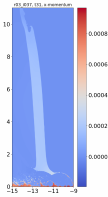} 
\includegraphics[width=0.22\textwidth]{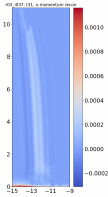} &
\includegraphics[width=0.22\textwidth]{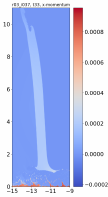} 
\includegraphics[width=0.22\textwidth]{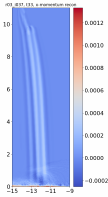} \\
\end{tabular}
\caption{Reconstruction of the variable {\it x-momentum}, after
  clustering, for test simulation {\bf r03\_i037} at time t31 (left)
  and t33 (right).  Top: the weight predictions (with 1 standard
  deviation error bar) for the first 50 weights. Bottom: each pair
  shows the plate view of the original snapshot and the reconstruction
  using 50 weights.  The color bars are different across images.
}
\label{fig:recon_clus_nxmom_i037}
\end{figure}


\begin{figure}
\centering
\begin{tabular}{cc}
\includegraphics[width=0.48\textwidth]{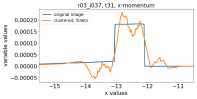} &
\includegraphics[width=0.48\textwidth]{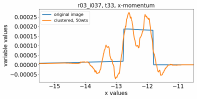} \\
\end{tabular}
\caption{Comparison of the reconstruction of the variable {\it
    x-momentum}, with clustering, for test simulation {\bf r03\_i037}
  at time t31 (left) and t33 (right).  The plots compare the value of
  the variable at $y= 6.0063$ for the reconstruction to
  indicate how well it detects the plate boundary.}
\label{fig:recon_lineout_nxmom_i037}
\end{figure}


\begin{figure}
\centering
\begin{tabular}{ccc}
\includegraphics[width=0.45\textwidth]{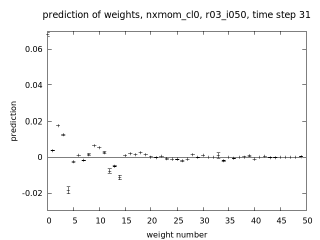} &
\includegraphics[width=0.45\textwidth]{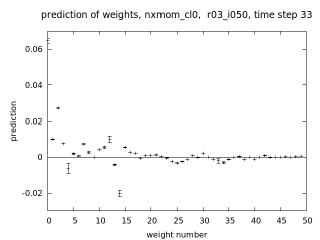} \\
\includegraphics[width=0.22\textwidth]{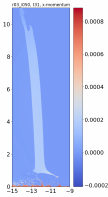} 
\includegraphics[width=0.22\textwidth]{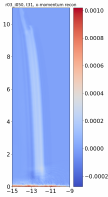} &
\includegraphics[width=0.22\textwidth]{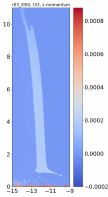} 
\includegraphics[width=0.22\textwidth]{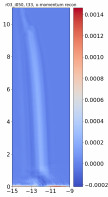} \\
\end{tabular}
\caption{Reconstruction of the variable {\it x-momentum}, after
  clustering, for test simulation {\bf r03\_i050} at time t31 (left)
  and t33 (right).  Top: the weight predictions (with 1 standard
  deviation error bar) for the first 50 weights. Bottom: each pair
  shows the plate view of the original snapshot and the reconstruction
  using 50 weights.  The color bars are different across images.}
\label{fig:recon_clus_nxmom_i050}
\end{figure}


\begin{figure}
\centering
\begin{tabular}{cc}
\includegraphics[width=0.48\textwidth]{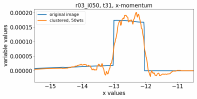} &
\includegraphics[width=0.48\textwidth]{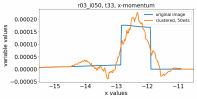} \\
\end{tabular}
\caption{Comparison of the reconstruction of the variable {\it
    x-momentum}, with clustering, for test simulation {\bf r03\_i050}
  at time t31 (left) and t33 (right).  The plots compare the value of
  the variable at $y= 6.0063$ for the reconstruction to
  indicate how well it detects the plate boundary.}
\label{fig:recon_lineout_nxmom_i050}
\end{figure}

\clearpage

\end{document}